\documentclass{article}
\usepackage[preprint]{neurips_2026}

\usepackage[utf8]{inputenc}
\usepackage[T1]{fontenc}
\usepackage{hyperref}
\usepackage{url}
\usepackage{booktabs}
\usepackage{amsfonts}
\usepackage{nicefrac}
\usepackage{microtype}
\usepackage{xcolor}

\usepackage{amsmath, amssymb}
\usepackage{graphicx}
\usepackage{float}
\usepackage{capt-of}  
\usepackage{wrapfig}  
\usepackage{multirow}
\usepackage[breakable]{tcolorbox}  
\usepackage{pifont}  
\usepackage{arydshln}  
\usepackage{colortbl}  
\usepackage{enumitem}
\usepackage{marvosym}
\setlist[itemize]{leftmargin=2em}
\setlist[enumerate]{leftmargin=2em}

\definecolor{darkblue}{rgb}{0, 0, 0.5}
\hypersetup{colorlinks=true, citecolor=darkblue, linkcolor=darkblue, urlcolor=darkblue}

\newcommand{\xmltag}[1]{\textcolor{teal}{\texttt{#1}}}
\newcommand{\phead}[1]{\textcolor{violet}{\texttt{#1}}}

\title{Where, What, Why, and Importance: Structured Defect Grounding for Text-to-Image Feedback}

\author{%
  \textbf{Huaisong Zhang\textsuperscript{1,2,*,\S} \quad Hao Yu\textsuperscript{1,*} \quad Yuxuan Zhang\textsuperscript{2,3,4,\S} \quad Jiahe Wang\textsuperscript{1} \quad Xinrui Chen\textsuperscript{1}} \\
  \textbf{Haoxiang Cao\textsuperscript{2,5,\S} \quad Feng Lu\textsuperscript{1} \quad Wendong Zhang\textsuperscript{2,\textdagger} \quad Changqian Yu\textsuperscript{2,\textdaggerdbl} \quad Chun Yuan\textsuperscript{1,\textdagger}} \\
  \parbox{0.95\textwidth}{\centering
  \textsuperscript{1}Tsinghua University \quad
  \textsuperscript{2}Kolors Team, KlingAI Research \\
  \textsuperscript{3}University of British Columbia \quad
  \textsuperscript{4}Vector Institute \quad
  \textsuperscript{5}South China Normal University
  }
}

\begin{document}

\maketitle
\begingroup
\renewcommand{\thefootnote}{}
\footnotetext{\parbox[t]{\dimexpr\linewidth-1.8em\relax}{\textsuperscript{*}Equal contribution. \quad \textsuperscript{\textdaggerdbl}Project lead. \quad \textsuperscript{\textdagger}Corresponding authors.\\\textsuperscript{\S}Work conducted during an internship at Kolors Team, KlingAI Research.}}
\endgroup


\begin{abstract}
Despite generating increasingly photorealistic images, text-to-image (T2I) models still exhibit localized, subtle, and structurally complex failures.
Diagnosing these failures requires instance-level feedback that answers \textit{where} a defect occurs, \textit{what} type it is, \textit{why} it is defective, and its \textit{importance} to overall image quality.
While recent dense-feedback methods move beyond scalar supervision, their heatmap-centric representations still formulate diagnosis as pixel-field regression, making it difficult to localize variable-cardinality defects and bind semantic reasons to individual failures. 
To address this representation bottleneck, we propose \textbf{Structured Defect Grounding (SDG)}, which casts T2I diagnosis as structured set prediction by modeling each defect as a \emph{(location, type, reason, importance)} tuple. 
To make this formulation trainable and measurable, we introduce \textbf{SDG-30K}, a 30K-image dataset with box-grounded annotations across four modern T2I generators, together with a dedicated evaluation protocol, \textbf{SDG-Eval}. 
Building on this structured representation, we further present a diagnosis-to-alignment framework in which a Vision-Language Model (VLM) serves as the SDG detector, and \textbf{BoxFlow-GRPO} converts predicted defect sets into box-derived, importance-weighted spatial rewards for diffusion model alignment. 
Extensive experiments show that our SDG detector outperforms leading proprietary VLMs on structured defect grounding, while SDG-guided rewards consistently improve T2I alignment and support localized image refinement. 
These results establish SDG as a unified, instance-level interface for diagnosing, evaluating, and enhancing modern generative models.
\end{abstract}

\section{Introduction}

\begin{figure*}[t]
\centering
\includegraphics[width=\textwidth]{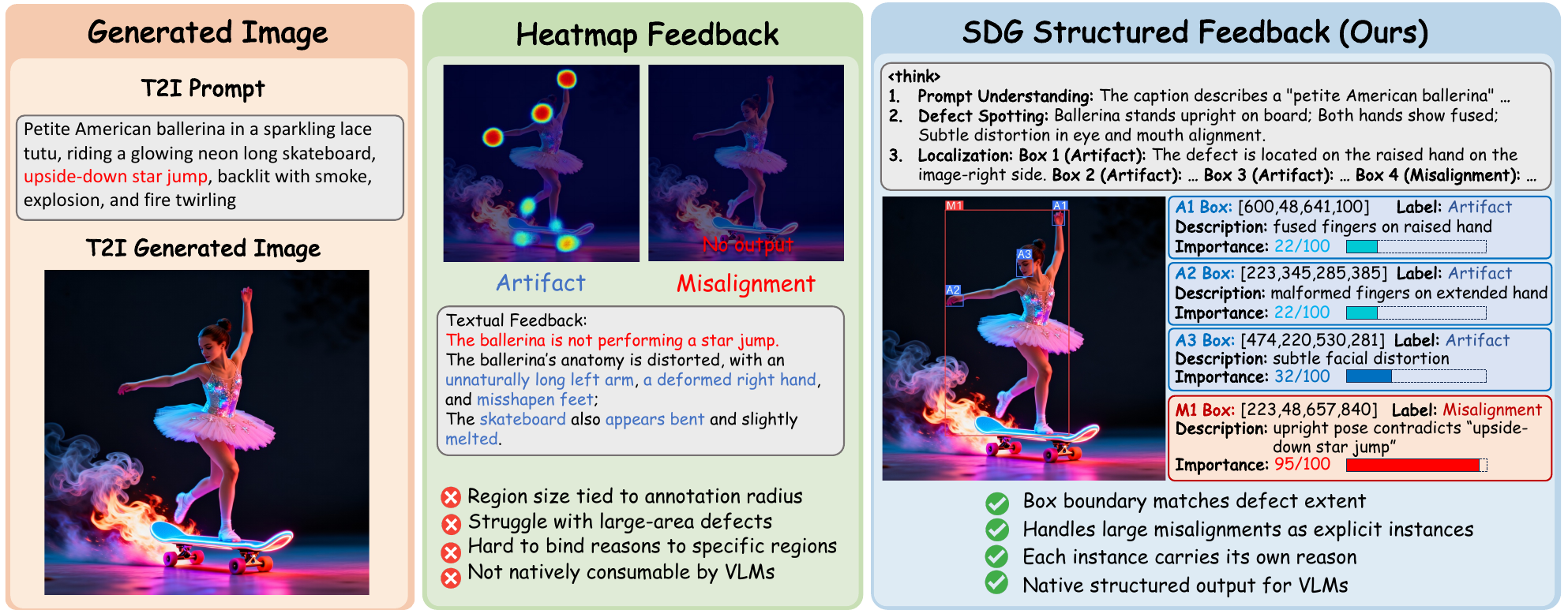}
\vspace{-18pt}
\caption{Comparison of dense feedback paradigms for T2I defect diagnosis. \textbf{Left}: a T2I generated image with its prompt. \textbf{Middle}: heatmap-style feedback produces separate artifact and misalignment maps with textual feedback. \textbf{Right}: SDG reasons via a chain-of-thought (CoT) trace, then outputs structured defects with bounding boxes, types, descriptions, and importance scores.}
\label{fig:heatmap_vs_defectset}
\vspace{-14pt}
\end{figure*}



Text-to-image (T2I) models~\citep{esser2024sd3,betker2023dalle3,flux2025,zimage2025,ma2025longcat,xie2025sana15} produce increasingly photorealistic images, yet their failures often remain localized, subtle, and structurally heterogeneous, such as 
malformed text, implausible geometry, and semantic mismatches.
Evaluating such outputs with scalar preference scores~\citep{wu2023humanpreferencescorebetter,wu2023humanpreferencescorev2,xu2023imagerewardlearningevaluatinghuman,kirstain2023pickapicopendatasetuser} collapses these defects into a single global value.
Although useful for ranking images, scalar feedback cannot answer the four key questions required for defect diagnosis: 
    \textit{where} the error occurs, 
    \textit{what} type of defect it is, 
    \textit{why} the region is defective, 
    and its \textit{importance} to overall image quality.
This limitation reveals a mismatch between global image-level supervision and the localized, instance-level nature of T2I failures.

Recent work attempts to address this need by moving beyond scalar supervision toward dense feedback. RichHF~\citep{liang2024richhumanfeedbacktexttoimage} introduces artifact and misalignment heatmaps, while ImageDoctor~\citep{guo2025imagedoctordiagnosingtexttoimagegeneration} augments heatmap-based diagnosis with VLM reasoning. 
Although heatmaps provide dense signals, they still formulate defect diagnosis as pixel-field regression rather than instance-level understanding. As illustrated in Figure~\ref{fig:heatmap_vs_defectset}, this heatmap-centric paradigm exposes representation bottlenecks: 
    spatially, point-derived maps depend on annotator-chosen radii rather than true defect extents~\citep{liang2024richhumanfeedbacktexttoimage}; 
    semantically, continuous severity fields cannot bind defect types, reasons, or importance scores to individual failures; 
    and architecturally, pixel-level maps are not native outputs of autoregressive VLMs and often require additional decoders or regression heads. 
These limitations suggest that the key challenge is not merely to make feedback dense, but to represent defects as spatially grounded, semantically explicit, and VLM-compatible instances.


Motivated by this bottleneck, we move from continuous pixel-field regression to structured instance prediction.
We introduce \textbf{Structured Defect Grounding (SDG)}, which models each defect as a structured \emph{(location, type, reason, importance)} tuple and formulates diagnosis as predicting a variable-length set of such tuples. 
SDG unifies two defect types, \emph{artifacts} (i.e., image-intrinsic visual flaws) and \emph{misalignments} (i.e., prompt-conditioned semantic errors), within a single instance space, naturally supporting images with multiple heterogeneous defects.
To make this formulation trainable and measurable, we construct \textbf{SDG-30K}, a 30,096-image dataset with box-grounded artifact and misalignment annotations across four modern T2I generators, and define \textbf{SDG-Eval}, a dedicated protocol for structured defect-set evaluation.



Beyond diagnosis, SDG also provides a natural interface for model alignment: boxes define spatial support, defect types and reasons provide semantic diagnoses, and importance scores calibrate reward strength. We convert SDG predictions into box-derived, importance-weighted spatial reward maps and post-train diffusion models with our proposed \textbf{BoxFlow-GRPO}, enabling spatially targeted alignment beyond scalar preference optimization. Code and released assets are available online.\footnote{\url{https://github.com/nianbai006/SDG}} See Appendix~\ref{app:data_code_availability} for details.
Overall, this paper makes following main \textbf{contributions}:



\begin{itemize}
\item We introduce \textbf{Structured Defect Grounding (SDG)}, an instance-level representation that formulates dense T2I diagnosis as variable-cardinality set prediction over structured \emph{(location, type, reason, importance)} tuples.
\item We construct \textbf{SDG-30K}, a 30,096-image box-grounded defect dataset across four modern T2I generators, and define \textbf{SDG-Eval} for image-level and defect-level evaluation.
\item We develop a diagnosis-to-alignment framework where a VLM-based \textbf{SDG detector} predicts structured defect sets and \textbf{BoxFlow-GRPO} converts them into importance-weighted spatial rewards for diffusion model alignment.
\item Extensive experiments demonstrate that our SDG detector outperforms leading proprietary VLMs on structured defect grounding, while SDG-guided rewards improve T2I alignment and support more faithful, actionable image refinement.
\end{itemize}


\section{Related Work}

\textbf{From scalar evaluation to dense T2I feedback.}
\label{sec:rw_dense}
Most T2I evaluators produce scalar scores~\citep{wu2023humanpreferencescorebetter,wu2023humanpreferencescorev2,xu2023imagerewardlearningevaluatinghuman,kirstain2023pickapicopendatasetuser}. RichHF~\citep{liang2024richhumanfeedbacktexttoimage} introduces heatmap-based dense feedback, ImageDoctor~\citep{guo2025imagedoctordiagnosingtexttoimagegeneration} predicts heatmaps via a VLM-plus-decoder architecture, and HEIE~\citep{yang2025heiemllmbasedhierarchicalexplainable} and MagicMirror~\citep{wang2025magicmirrorlargescaledatasetbenchmark} further enriches feedback with hierarchical explanations and fine-grained artifact taxonomy, respectively. Beyond T2I evaluation, HAD~\citep{wang2025detectinghumanartifactstexttoimage} and AbHuman~\citep{fang2024humanrefinerbenchmarkingabnormalhuman} demonstrate box-level supervision for human-body artifact detection, and LEGION~\citep{kang2025legionlearninggroundexplain} combines mask-level localization with natural-language explanations. 
However, they do not provide a unified instance-level formulation that jointly localizes \textit{artifact} and \textit{misalignment} with localized descriptions.

\textbf{Structured spatial reasoning in VLMs.}
Modern VLMs increasingly support explicit spatial structure within an autoregressive generation framework. Qwen2.5-VL~\citep{qwen2025qwen25vl} supports box- and point-based grounding with structured outputs, while Qwen3-VL~\citep{qwen2025qwen3vl} further strengthens image-grounded reasoning and spatial understanding. At a finer granularity, SimpleSeg~\citep{song2026pixellevelvlmperceptionsimple} reformulates segmentation as point-sequence generation entirely in language space. These advances motivate our use of VLMs to generate structured defect instances.

\textbf{RL for diffusion alignment and image refinement.}
RL has become a key tool for aligning diffusion models with human preferences. DDPO~\citep{black2024ddpo} frames denoising as a multi-step MDP and applies policy gradients, Diffusion-DPO~\citep{wallace2024diffusiondpo} adapts direct preference optimization to diffusion training, and Flow-GRPO~\citep{zhang2025flowgrpo} applies group relative policy optimization~\citep{shao2024deepseekmath} to flow-matching models. Related work adapts diffusion RL across denoising steps through entropy regulation~\citep{yan2025entropy} and stage-specific reward objectives~\citep{yan2026explore}. TempFlow-GRPO~\citep{he2025tempflowgrpo} uses trajectory branching and noise-aware timestep weighting to assign credit across denoising steps. ImageDoctor~\citep{guo2025imagedoctordiagnosingtexttoimagegeneration} proposes extending scalar rewards to spatially varying dense reward maps. On the correction side, ReflectionFlow~\citep{zhuo2025reflectionflow} and HumanRefiner~\citep{fang2024humanrefinerbenchmarkingabnormalhuman} demonstrate that defect-guided refinement can improve generation quality. To our knowledge, our work is the first to realize spatially dense advantages in diffusion RL, and further uses structured dense feedback to guide image refinement.

\section{SDG-30K: Dataset and Evaluation}
\label{sec:dataset}

\subsection{Task Formulation}
\label{sec:formulation}

Given a generated image $I$ and its prompt $T$, the goal is to predict a variable-cardinality set of structured defect instances $\{(b_i, t_i, r_i, s_i)\}_{i=1}^{K}$.
Here, $K \in \mathbb{N}$ denotes the number of defects in the image. 
The $i$-th instance consists of
(1) a quantized bounding box $b_i \in \{0,\dots,1000\}^4$; 
(2) a defect type $t_i \in \{\textit{artifact},\, \textit{misalignment}\}$; 
(3) a free-form natural-language description $r_i$; and 
(4) an integer importance score $s_i \in \{1,\dots,100\}$ reflecting the defect's perceptual impact on the overall image quality.
The detailed definitions and distinctions of these defect types are provided in Appendix~\ref{app:annotation-guideline}.

\begin{figure*}[t]
\centering
\includegraphics[width=\textwidth]{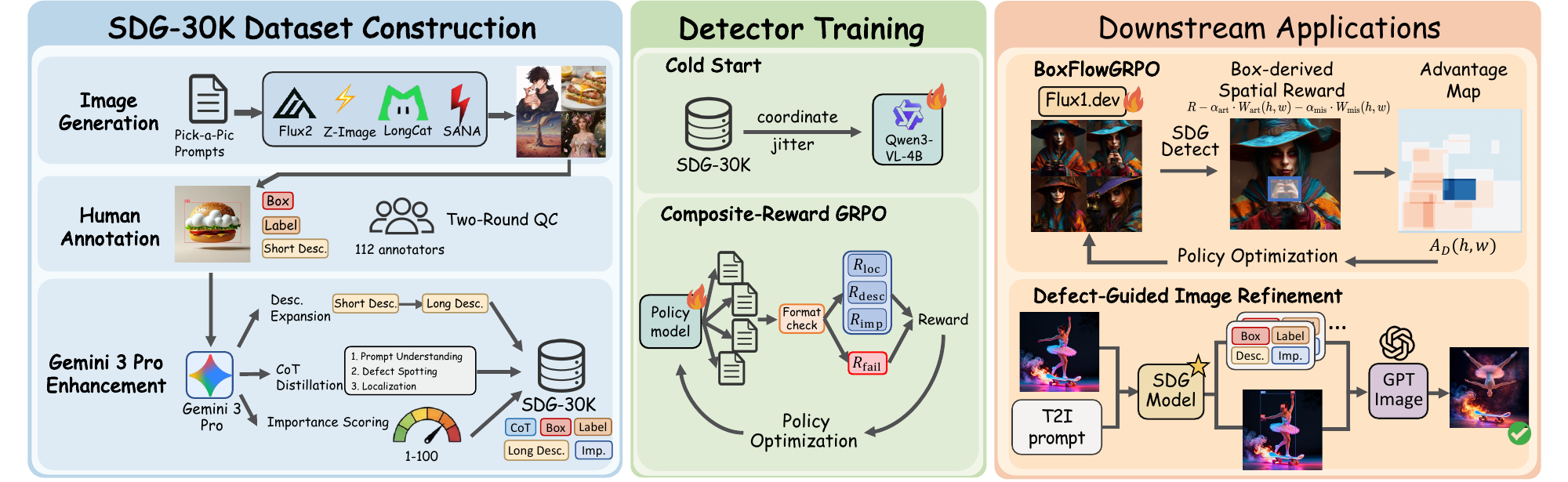}
\vspace{-18pt}
\caption{Overview of the SDG framework. \textbf{Left}: SDG-30K construction combines human box-level defect annotation across four T2I generators with Gemini 3 Pro enhancement. \textbf{Middle}: Two-stage SDG detector training via SFT with coordinate jitter followed by GRPO with a format-gated composite reward. \textbf{Right}: Downstream applications. BoxFlow-GRPO converts SDG detections into box-derived spatial rewards for diffusion alignment; defect-guided refinement feeds box overlays and text feedback to GPT-Image-1.5 for image refinement.}
\label{fig:pipeline}
\vspace{-14pt}
\end{figure*}



\subsection{Evaluation Metrics}
\label{sec:eval_protocol}

We define \textbf{SDG-Eval}, an evaluation protocol that scores SDG at both image and defect levels for each defect type $t \in \{\textit{artifact}, \textit{misalignment}\}$; full definitions are provided in Appendix~\ref{app:eval_metrics}. At the image level, \textbf{DetTypeF1} measures whether a defect type is present on each test image, while \textbf{ClnAcc} reports true-negative accuracy on images with no ground-truth instance of that type ($GT{=}0$). At the defect level, predictions are matched to ground-truth instances of the same type via class-aware Hungarian matching by IoU. We report \textbf{BoxF1}@0.1 and \textbf{BoxF1}@0.5 for localization, \textbf{DescCos}@0.1 using Qwen3-Embedding-0.6B for description alignment, and \textbf{ImpAcc}@0.1 as normalized importance accuracy over matched pairs.

\subsection{Dataset Construction}
\label{sec:construction}

Figure~\ref{fig:pipeline} (left) illustrates the dataset construction pipeline. SDG-30K contains \textbf{30,096} images at $1024{\times}1024$ resolution, generated from Pick-a-Pic prompts~\citep{kirstain2023pickapicopendatasetuser} using four T2I generators (${\sim}$7.5K each): FLUX.2-dev~\citep{flux2025}, Z-Image-Turbo~\citep{zimage2025}, LongCat-Image~\citep{ma2025longcat}, and SANA-1.5-1.6B~\citep{xie2025sana15}.

\textbf{112} annotators (${\sim}$\textbf{1,085} person-hours) examined each prompt--image pair, drew bounding boxes, assigned top-level labels, and wrote concise Chinese descriptions ($\le$30 characters). Two rounds of review resolved disagreements. To quantify inter-annotator agreement, 16 held-out annotators who did not participate in the original labeling independently re-annotated the test set. The resulting BoxF1@0.5 of 0.278 (\textit{artifact}) and 0.409 (\textit{misalignment}) against the primary annotations serves as a human upper bound for localization (Table~\ref{tab:main_art_mis_results}).
Detailed annotation rules are in Appendix~\ref{app:annotation-guideline}.

Human annotations provide boxes, labels, and short Chinese descriptions but lack reasoning traces and importance scores. We use Gemini 3 Pro~\citep{geminiteam2025geminifamilyhighlycapable} to augment each sample with three components: (1)~\emph{description expansion}:  from Chinese to detailed English; (2)~\emph{reasoning trace distillation}: as a three-step CoT trace (prompt understanding, defect spotting, localization); and (3)~\emph{importance scoring}: based on a multi-criteria rubric. Details are in Appendix~\ref{app:gemini_prompt}. Human calibration on 500 distinct defect boxes yields 64.8\% exact-tier agreement, 97.2\% agreement within one adjacent tier, and Spearman correlation 0.786 with Gemini scores (Appendix~\ref{app:importance_human_validation}). Table~\ref{tab:dataset_comparison} compares SDG-30K with related datasets.

\begin{table*}[t]
\centering
\caption{Comparison with related datasets. SDG-30K is the first to jointly cover \textit{artifact} and \textit{misalignment} within a unified instance-level annotation space with natural-language reasons.}
\label{tab:dataset_comparison}
\small
\setlength{\tabcolsep}{5pt}
\resizebox{\textwidth}{!}{%
\begin{tabular}{lccccccc}
\toprule
Dataset & Images & Spatial Ann. & Defect Types & NL Reason & Instance-Level & Generators \\
\midrule
RichHF-18K~\citep{liang2024richhumanfeedbacktexttoimage} & 18K & Heatmap (point-derived) & Art.\ + Mis.\ & $\times$ & $\times$ & SD-family \\
HAD~\citep{wang2025detectinghumanartifactstexttoimage} & 37K & Bounding box & Artifact (human body) & $\times$ & $\checkmark$ & SDXL, DALL-E, MJ \\
AbHuman~\citep{fang2024humanrefinerbenchmarkingabnormalhuman} & 56K & Bounding box & Artifact (human body) & $\times$ & $\checkmark$ & SDXL \\
MagicMirror~\citep{wang2025magicmirrorlargescaledatasetbenchmark} & 343K & Taxonomy label & Artifact & $\times$ & $\times$ & FLUX, SD3, Kolors, MJ, etc. \\
LEGION~\citep{kang2025legionlearninggroundexplain} & 12K & Pixel mask & Artifact & $\checkmark$ & $\checkmark$ & SD, FLUX, GLIDE, etc. \\
\midrule
\textbf{SDG-30K (Ours)} & 30K & Bounding box & Art.\ + Mis.\  & $\checkmark$ & $\checkmark$ & FLUX, Z-Image, LongCat, SANA \\
\bottomrule
\end{tabular}%
}
\vspace{-4pt}
\includegraphics[width=\textwidth]{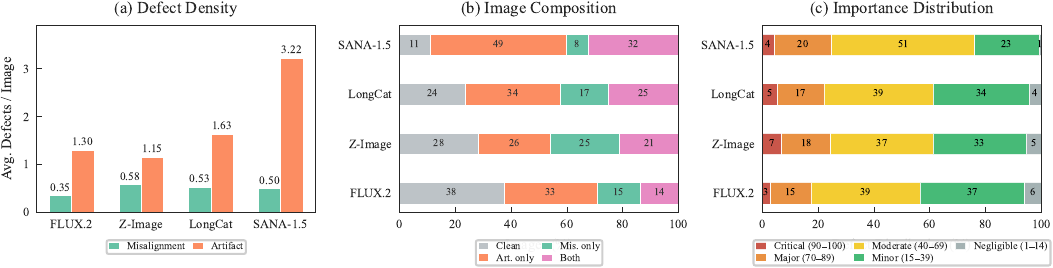}
\vspace{-14pt}
\captionof{figure}{SDG-30K dataset statistics across four generators. (a) Average number of defects per image by type. (b) Image composition showing the proportion of clean, artifact-only, misalignment-only, and both-type images. (c) Distribution of defect importance scores across five severity tiers.}
\label{fig:dataset_stats}
\vspace{-12pt}
\end{table*}

\subsection{Statistics and Splits}

\textbf{Dataset split.}
The split follows the Pick-a-Pic prompt partition (prompt-disjoint): \textbf{28,945} training and \textbf{1,151} test images, ensuring that no prompt appears in both splits.

\textbf{Defect frequency and image composition.}
Figure~\ref{fig:dataset_stats} summarizes the dataset statistics. \textit{Artifact} instances are more frequent than \textit{misalignment} across all four generators, with SANA-1.5 exhibiting the highest artifact frequency (3.22 per image). Weighting the per-generator proportions in Figure~\ref{fig:dataset_stats}b by image count, approximately 25.1\% of images are defect-free, 35.5\% contain only artifacts, 16.3\% contain only misalignments, and 23.1\% contain both. This confirms that localized failures remain common in recent T2I generators, motivating dense, instance-level feedback.

\textbf{Importance distribution.}
Importance scores estimate each defect's impact on image quality and prompt faithfulness, enabling failure prioritization and reward weighting. They are centered around the Moderate tier (40--69), mostly between 30 and 80 (Figure~\ref{fig:dataset_stats}c), with rare extremes and balanced coverage for training.

\section{Structured Defect Grounding Framework}
\label{sec:method}

In this section, we instantiate SDG as a diagnosis-to-alignment framework built around the structured-defect representation. We first present the SDG detector, which produces structured defect sets that inherently align with VLM output formats. We then introduce BoxFlow-GRPO to convert SDG's outputs into box-derived spatial rewards for diffusion alignment.


\subsection{SDG Detector}
\label{sec:training}




We formulate defect diagnosis as a structured vision-language generation task. Given a generated image and its prompt, SDG first produces a reasoning trace $\mathcal{R}$ and then emits a structured defect set $\mathcal{D}$, where each instance specifies its location, type, reason, and importance.

The SDG detector is trained in two stages (Figure~\ref{fig:pipeline}, middle). Supervised fine-tuning (SFT) teaches the model to follow the defect-grounding instruction and emit the required structured format, while group relative policy optimization (GRPO)~\citep{shao2024deepseekmath} further improves localization, description consistency, and importance estimation under a format-validity gate.

\textbf{Cold Start. }We fine-tune Qwen3-VL-4B-Instruct on the SDG-30K training split (Section~\ref{sec:construction}). For each image-prompt pair $(I,T)$, the target sequence concatenates the Gemini-distilled reasoning trace and the structured defect set, denoted as $y=[\mathcal{R};\mathcal{D}]$. We optimize the model with the standard teacher-forced negative log-likelihood over the target tokens:
\begin{equation}
\mathcal{L}_{\mathrm{SFT}} = -\sum_{t=1}^{|y|} \log \pi_\theta(y_t \mid I, T, y_{<t}).
\end{equation}

To reduce sensitivity to exact coordinate values, we apply \emph{coordinate jitter} to the \texttt{<answer>} segment when preparing the SFT data. For each box $[x_0, y_0, x_1, y_1]$ in normalized $[0,1000]$ space, each coordinate is independently perturbed by $\delta \sim \mathcal{U}(-10,10)$, then clamped to $[0,1000]$ with valid box ordering enforced. The \texttt{<think>} segment is kept unchanged because it captures high-level visual reasoning rather than exact numerical coordinates. With independently sampled jitter offsets for the repeated training examples, the model observes multiple plausible coordinate variants of the same defect, improving tolerance to minor spatial variation and providing a more robust SFT initialization for subsequent GRPO training.

\textbf{Composite-Reward GRPO. }After SFT, we apply GRPO to directly optimize the structured output using rewards for spatial accuracy, description consistency, and importance estimation. For each prompt, we sample $S{=}8$ responses, compute the composite reward $R_s$ for each response, and form group-normalized advantages $A_s=(R_s-\bar{R})/\sigma_R$. The policy is then optimized with the clipped GRPO objective:
\begin{equation}
\mathcal{L}_{\mathrm{GRPO}} =
-\mathbb{E}\!\left[
\min\!\left(
\rho_s A_s,\;
\mathrm{clip}(\rho_s, 1{-}\epsilon, 1{+}\epsilon)\, A_s
\right)
\right]
+ \beta\, \mathrm{KL}(\pi_\theta \,\|\, \pi_{\mathrm{ref}}),
\end{equation}
where $\rho_s = \pi_\theta(y_s \mid I,T) / \pi_{\mathrm{old}}(y_s \mid I,T)$ is the importance ratio, $\pi_{\mathrm{old}}$ is the rollout policy, $\pi_{\mathrm{ref}}$ is the fixed reference policy, and $\beta{=}0.01$.

The composite reward is gated by a format-validity check:
\begin{equation}
R = \begin{cases}
\lambda_{\mathrm{loc}}\, R_{\mathrm{loc}} + \lambda_{\mathrm{desc}}\, R_{\mathrm{desc}} + \lambda_{\mathrm{imp}}\, R_{\mathrm{imp}}, & \text{if } \mathrm{Format}(y) = \mathrm{true}, \\[2pt]
R_{\mathrm{fail}}, & \text{otherwise},
\end{cases}
\end{equation}
where $\lambda_{\mathrm{loc}}, \lambda_{\mathrm{desc}}, \lambda_{\mathrm{imp}} \ge 0$ with $\lambda_{\mathrm{loc}} + \lambda_{\mathrm{desc}} + \lambda_{\mathrm{imp}} = 1$ control the relative importance of each reward component, and $R_{\mathrm{fail}} < 0$ is a fixed penalty for malformed outputs. The format predicate $\mathrm{Format}(y)$ verifies that the response contains well-formed reasoning and answer delimiters, a parseable JSON defect list, and geometrically valid bounding boxes (i.e., $x_0 < x_1$ and $y_0 < y_1$).

The three reward components are defined as follows. $R_{\mathrm{loc}}$ measures spatial grounding accuracy: predicted and ground-truth boxes are matched via the Hungarian algorithm with Distance-IoU (DIoU) cost, and the reward reflects match quality with explicit penalties for false negatives and false positives. $R_{\mathrm{desc}}$ measures description consistency via embedding cosine similarity (Qwen3-Embedding-0.6B), linearly mapped to $[0,1]$ and averaged over matched pairs. $R_{\mathrm{imp}}$ measures importance estimation accuracy via a clipped absolute-error metric. In practice, we set $\lambda_{\mathrm{loc}}{=}0.6$, $\lambda_{\mathrm{desc}}{=}0.25$, $\lambda_{\mathrm{imp}}{=}0.15$, and $R_{\mathrm{fail}}{=}{-1}$. Detailed formulas and the treatment of boundary cases are provided in Appendix~\ref{app:experimental_details}.

\subsection{SDG-Guided BoxFlow-GRPO}
\label{sec:denseflow}




Once the SDG detector is trained, we translate its dense feedback into spatially varying rewards to guide the reinforcement learning of diffusion models. 
The closest prior work is ImageDoctor's DenseFlow-GRPO~\citep{guo2025imagedoctordiagnosingtexttoimagegeneration}, which extends Flow-GRPO~\citep{zhang2025flowgrpo} with a heatmap signal. In its public training code\footnote{\url{https://github.com/EthanG97/ImageDoctor}}, each image is assigned an image-level scalar reward $R$, normalized across the prompt group into a scalar advantage $A$; the predicted heatmap $H \in [0,1]^{H \times W}$ enters only as a multiplicative mask on the policy-gradient loss, $\mathcal{L} \propto -A \cdot \rho \cdot (1{-}H)$, where $\rho$ is the importance ratio. The gradient signal is therefore still driven by an image-level scalar, with the heatmap merely down-weighting locations the detector flags as defective rather than contributing a true per-location advantage at the latent grid.

Motivated by this gap, we implement BoxFlow-GRPO around two design choices: box-derived spatial rewards and spatially normalized per-location advantages.


Given a base scalar reward $R$ from any predefined reward model\footnote{We use UnifiedReward-2.0 (UR2)~\citep{wang2026unifiedrewardmodelmultimodal} in our experiments} and the defect bounding boxes detected by our SDG model, we construct a spatially varying reward map in the latent space. For each latent spatial location $(h,w)$, weighted masks $W_{\mathrm{art}}, W_{\mathrm{mis}}$ aggregate the predicted importance of all defect boxes covering that location, and the per-location reward subtracts type-specific penalties from the scalar reward:
\begin{equation}
\begin{gathered}
W_{\mathrm{type}}(h,w) = \max_{k \in \mathcal{B}_{\mathrm{type}}(h,w)} \hat{s}_k / 100, \\
\alpha_{\mathrm{art}} = c_{\mathrm{art}} \cdot \sigma_R^{(\mathrm{group})}, \quad
\alpha_{\mathrm{mis}} = c_{\mathrm{mis}} \cdot \sigma_R^{(\mathrm{group})}, \\
R_D(h,w) = R - \alpha_{\mathrm{art}} W_{\mathrm{art}}(h,w) - \alpha_{\mathrm{mis}} W_{\mathrm{mis}}(h,w).
\end{gathered}
\end{equation}
where $\mathcal{B}_{\mathrm{type}}(h,w)$ denotes the set of boxes of a given type covering location $(h,w)$ and $\hat{s}_k \in \{1, \dots, 100\}$ is the importance predicted by SDG for box $k$ (Section~\ref{sec:eval_protocol}). We set $W_{\mathrm{type}}(h,w)=0$ when no box of that type covers $(h,w)$.
We set $c_{\mathrm{art}}{=}0.5$ and $c_{\mathrm{mis}}{=}0.05$ to approximately normalize spatial penalty mass: an average misalignment box covers $9.90\times$ the area of an artifact box. Among three coefficient settings evaluated after 500 training steps, only this setting improves all five metrics relative to Base (Appendix~\ref{app:coefficient_sensitivity}). Scaling by $\sigma_R^{(\mathrm{group})}$ adapts penalties to reward variability, while importance weights prioritize severe defects.


After constructing $R_D$, per-location advantages are computed by normalizing across the $K$ samples within each prompt group at each spatial location:
\begin{equation}
A_D^{(k)}(h,w) = \frac{R_D^{(k)}(h,w) - \mu_D(h,w)}{\sigma_D(h,w) + \epsilon},
\end{equation}
\vspace{-8pt}

where $\mu_D(h,w)$ and $\sigma_D(h,w)$ are the mean and standard deviation of $\{R_D^{(k)}(h,w)\}_{k=1}^{K}$ computed over the $K{=}8$ samples in the prompt group, and $\epsilon$ is a small constant for numerical stability. Defining the per-location likelihood ratio $\rho_t^{(k)}(h,w) = \pi_\phi(x_{t-1,h,w}^{(k)} \mid x_t^{(k)}, c) / \pi_{\phi_{\mathrm{old}}}(x_{t-1,h,w}^{(k)} \mid x_t^{(k)}, c)$, the BoxFlow-GRPO objective is (KL regularization omitted for brevity):
\begin{equation}
\resizebox{\linewidth}{!}{$\displaystyle
\mathcal{J}_{\mathrm{BoxFlow}}(\phi) = \frac{1}{KTHW} \sum_{k,t,h,w} \min\!\left( \rho_t^{(k)}(h,w)\, A_D^{(k)}(h,w),\; \mathrm{clip}\!\left(\rho_t^{(k)}(h,w), 1{-}\varepsilon, 1{+}\varepsilon\right) A_D^{(k)}(h,w) \right)
$}
\end{equation}
where $T$ is the number of denoising steps and $H, W$ are the latent spatial dimensions. Compared with a scalar-advantage implementation, this objective preserves spatial variation in both the advantage and the likelihood ratio.

\section{Experiments}
\label{sec:experiments}

We evaluate SDG from three complementary perspectives: (1)~defect grounding quality on SDG-30K, (2)~diffusion alignment with structured-feedback rewards via BoxFlow-GRPO, and (3)~defect-guided image refinement driven by structured feedback.

\subsection{Defect Grounding Results}
\label{sec:main_results}

\subsubsection{Setup}

\textbf{Implementation.}
We fine-tune Qwen3-VL-4B-Instruct on $16{\times}$A100-80G GPUs with DeepSpeed ZeRO-2 for one epoch over a $3{\times}$ pre-augmented SFT corpus (effective batch size 16, learning rate $3{\times}10^{-5}$, cosine schedule). The vision encoder is frozen throughout SFT; Table~\ref{tab:ablation_summary} shows that unfreezing it degrades grounding quality at this dataset scale. For GRPO, we train on $2{\times}8$ A100 GPUs (16 total) with $S{=}8$ sampled responses per prompt (temperature 1.0, top-$p$ 0.85) for 2 epochs at learning rate $5{\times}10^{-6}$.

\textbf{Baselines.}
We compare \textbf{SDG} (SFT and GRPO variants) against zero-shot GPT-5.4 and Gemini 3 Pro, both prompted with the same structured output format but without task-specific training. A human reference from 16 independent re-annotators provides a localization upper bound.

\subsubsection{Main Results}

\begin{table*}[h]
\vspace{-4pt}
\centering
\caption{Defect grounding results on SDG-30K test set. Human row: localization upper bound from 16 independent re-annotators. \textbf{Bold}: best; \underline{underline}: second best.}
\label{tab:main_art_mis_results}
\small
\setlength{\tabcolsep}{3pt}
\renewcommand{\arraystretch}{1.15}
\resizebox{\linewidth}{!}{%
\begin{tabular}{l|cccccc|cccccc}
\toprule
& \multicolumn{6}{c|}{\emph{Artifact}} & \multicolumn{6}{c}{\emph{Misalignment}} \\
\cmidrule(lr){2-7}\cmidrule(lr){8-13}
& \multicolumn{2}{c}{Image-Level} & \multicolumn{4}{c|}{Defect-Level} & \multicolumn{2}{c}{Image-Level} & \multicolumn{4}{c}{Defect-Level} \\
\cmidrule(lr){2-3}\cmidrule(lr){4-7}\cmidrule(lr){8-9}\cmidrule(lr){10-13}
& All & $GT{=}0$ & \multicolumn{4}{c|}{$GT{>}0$} & All & $GT{=}0$ & \multicolumn{4}{c}{$GT{>}0$} \\
Model & DetTypeF1 & ClnAcc & BoxF1@0.1 & BoxF1@0.5 & DescCos@0.1 & ImpAcc@0.1 & DetTypeF1 & ClnAcc & BoxF1@0.1 & BoxF1@0.5 & DescCos@0.1 & ImpAcc@0.1 \\
\midrule
GPT-5.4 & 0.736 & 0.118 & 0.237 & 0.035 & 0.868 & 0.812 & 0.654 & 0.513 & 0.457 & 0.292 & 0.837 & 0.820 \\
Gemini 3 Pro & 0.733 & 0.076 & 0.337 & 0.200 & \underline{0.885} & 0.824 & \textbf{0.696} & 0.672 & 0.464 & 0.307 & \underline{0.891} & 0.851 \\
\textbf{SDG (SFT)} & \textbf{0.776} & \textbf{0.697} & \underline{0.402} & \underline{0.255} & \textbf{0.904} & \underline{0.883} & 0.636 & \textbf{0.799} & \underline{0.499} & \underline{0.376} & \textbf{0.893} & \underline{0.892} \\
\textbf{SDG (GRPO)} & \underline{0.772} & \underline{0.560} & \textbf{0.404} & \textbf{0.263} & \textbf{0.904} & \textbf{0.887} & \underline{0.675} & \underline{0.732} & \textbf{0.511} & \textbf{0.387} & 0.888 & \textbf{0.893} \\
\cdashline{1-13}
\rowcolor{gray!10} Human & 0.792 & 0.774 & 0.462 & 0.278 & -- & -- & 0.712 & 0.818 & 0.563 & 0.409 & -- & -- \\
\bottomrule
\end{tabular}%
}
\vspace{-10pt}
\end{table*}

\textbf{Quantitative analysis.}
Table~\ref{tab:main_art_mis_results} presents defect grounding results for artifact and misalignment separately. GRPO achieves the strongest BoxF1@0.5 (0.263/0.387) and importance accuracy (0.887/0.893), while SFT remains competitive and attains the highest clean-image accuracy (0.697/0.799). Compared with zero-shot VLM baselines, both SDG variants substantially reduce localization error: GPT-5.4 obtains reasonable image-level artifact detection (F1 0.736) but weak precise artifact localization (BoxF1@0.5 0.035), and Gemini 3 Pro improves localization (0.200/0.307) but remains below SDG. GRPO narrows the gap to the human reference on both artifact (0.263 vs.\ 0.278) and misalignment (0.387 vs.\ 0.409), while maintaining high description cosine similarity.

\begin{figure}[t!]
\centering
\includegraphics[width=0.88\textwidth]{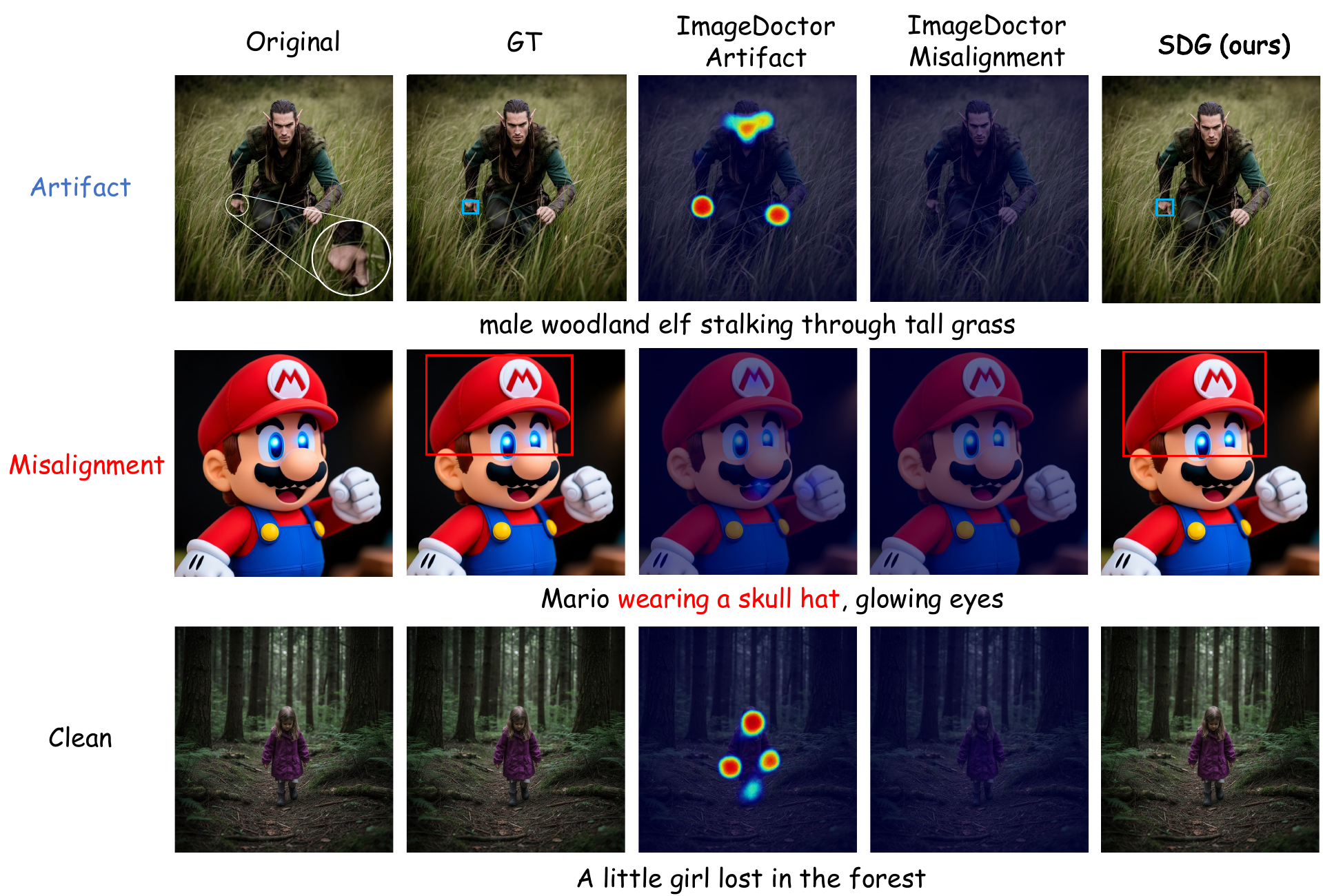}
\vspace{-8pt}
\caption{Qualitative comparison on SDG-30K. Rows from top to bottom: artifact-only, misalignment-only, and clean images.}
\label{fig:sdg_qualitative}
\vspace{-6pt}
\centering
\begin{minipage}[t]{0.56\textwidth}
\vspace{0pt}
\centering
\captionof{table}{Image-level defect detection on RichHF-18K. ImageDoctor is trained in-domain; SDG is zero-shot.}
\label{tab:richhf}
\footnotesize
\setlength{\tabcolsep}{3pt}
\renewcommand{\arraystretch}{1.12}
\resizebox{\linewidth}{!}{%
\begin{tabular}{l|ccc|ccc}
\toprule
& \multicolumn{3}{c|}{\emph{Artifact}} & \multicolumn{3}{c}{\emph{Misalignment}} \\
\cmidrule(lr){2-4}\cmidrule(lr){5-7}
Method & Prec. & Rec. & F1 & Prec. & Rec. & F1 \\
\midrule
ImageDoctor @0.10 & 0.965 & 0.939 & 0.952 & 0.983 & 0.143 & 0.250 \\
ImageDoctor @0.33 & 0.977 & 0.762 & 0.856 & 1.000 & 0.004 & 0.007 \\
\cdashline{1-7}
SDG (ours) & 0.970 & 0.698 & 0.812 & 0.951 & 0.500 & 0.655 \\
\bottomrule
\end{tabular}%
}
\end{minipage}
\vspace{-4pt}
\hfill
\begin{minipage}[t]{0.40\textwidth}
\vspace{0pt}
\centering
\captionof{table}{Key ablation results on SDG-30K. Full metrics are in Table~\ref{tab:ablation}.}
\label{tab:ablation_summary}
\footnotesize
\setlength{\tabcolsep}{4pt}
\renewcommand{\arraystretch}{1.12}
\resizebox{\linewidth}{!}{%
\begin{tabular}{lcc}
\toprule
Variant & Art. BoxF1@0.5 & Mis. BoxF1@0.5 \\
\midrule
SFT & 0.255 & 0.376 \\
GRPO & 0.263 & 0.387 \\
\cdashline{1-3}
SFT w/o step 1 & 0.218 & 0.329 \\
SFT w/o step 3 & 0.253 & 0.281 \\
GRPO w/o CoT & 0.288 & 0.352 \\
\cdashline{1-3}
SFT unfreeze ViT & 0.203 & 0.305 \\
SFT w/o jitter & 0.253 & 0.360 \\
\bottomrule
\end{tabular}%
}
\end{minipage}
\vspace{-4pt}

\vspace{-10pt}
\end{figure}

\textbf{Qualitative comparison.}
Figure~\ref{fig:sdg_qualitative} shows qualitative comparisons on SDG-30K across three representative cases: an \textit{artifact}-only image (top), a \textit{misalignment}-only image (middle), and a clean image (bottom). SDG produces instance-level bounding boxes with per-defect descriptions, accurately localizing defects in the first two rows and correctly recognizing the clean case as defect-free. In contrast, ImageDoctor tends to predict heatmap responses on faces and hands even when they are anatomically correct, and struggles to detect prompt-conditioned misalignments.

\textbf{Cross-dataset generalization.}
To evaluate generalization, we test SDG on the RichHF-18K test set without any fine-tuning on this dataset, comparing against ImageDoctor~\citep{guo2025imagedoctordiagnosingtexttoimagegeneration} which is trained on RichHF-18K. Since RichHF-18K provides heatmap annotations, we threshold ImageDoctor heatmaps at two operating points (0.10 and 0.33) for image-level evaluation.

As shown in Table~\ref{tab:richhf}, SDG achieves substantially higher misalignment F1 (0.655 vs.\ 0.250/0.007), demonstrating that structured defect grounding generalizes to unseen data and captures prompt-conditioned misalignments more effectively than heatmap-based methods. ImageDoctor achieves higher artifact F1 at the loose threshold (0.952), as expected from in-domain training, but its misalignment recall is notably poor (0.143/0.004), suggesting that heatmap-based representations struggle with prompt-conditioned misalignment in this setting.

\subsubsection{Ablation Study}

Table~\ref{tab:ablation_summary} summarizes the main ablation trends, with full metrics in Appendix~\ref{app:extended_results}. GRPO improves localization over SFT (artifact/misalignment BoxF1@0.5: 0.263/0.387 vs.\ 0.255/0.376), confirming that policy optimization refines spatial precision. Removing CoT steps hurts prompt-conditioned misalignment more than artifact detection, and removing the reasoning trace from GRPO lowers misalignment BoxF1@0.5 from 0.387 to 0.352. Unfreezing the vision encoder substantially degrades localization, while coordinate jitter mainly improves image-level robustness (full results in Table~\ref{tab:ablation}).

\subsection{Downstream Applications}
\label{sec:downstream}

\subsubsection{BoxFlow-GRPO}
\label{sec:denseflow_results}

\textbf{Setup.}
We apply the dense reward construction of Section~\ref{sec:denseflow} to FLUX.1-dev, following Flow-GRPO~\citep{zhang2025flowgrpo}. The scalar base reward is UnifiedReward-2.0 (UR2)~\citep{wang2026unifiedrewardmodelmultimodal}. We use the GRPO checkpoint of the SDG detector (Qwen3-VL-4B-Instruct) to identify defects and construct a dense reward by subtracting importance-weighted artifact and misalignment penalties at covered latent locations. Training uses Pick-a-Pic~\citep{kirstain2023pickapicopendatasetuser} prompts held out from SDG-30K (prompt-disjoint), for 500 optimization steps at $512{\times}512$ resolution on $8{\times}$A100 GPUs with LoRA (rank 64, $\alpha{=}128$) and learning rate $3{\times}10^{-4}$. We evaluate on DrawBench~\citep{saharia2022imagen} using PickScore~\citep{kirstain2023pickapicopendatasetuser}, CLIPScore~\citep{hessel2021clipscore}, HPSv3~\citep{ma2025hpsv3}, DeQA~\citep{you2025deqa}, and the real-image probability $P(\mathrm{real})$ from Forensic-Chat~\citep{lin2025seeingbeforereasoning}, obtained by a 2-class softmax over its ``real'' vs.\ ``fake'' token logits.

\begin{figure}[t]
\centering
\begin{minipage}[t]{0.49\textwidth}
\centering
\includegraphics[width=0.82\textwidth]{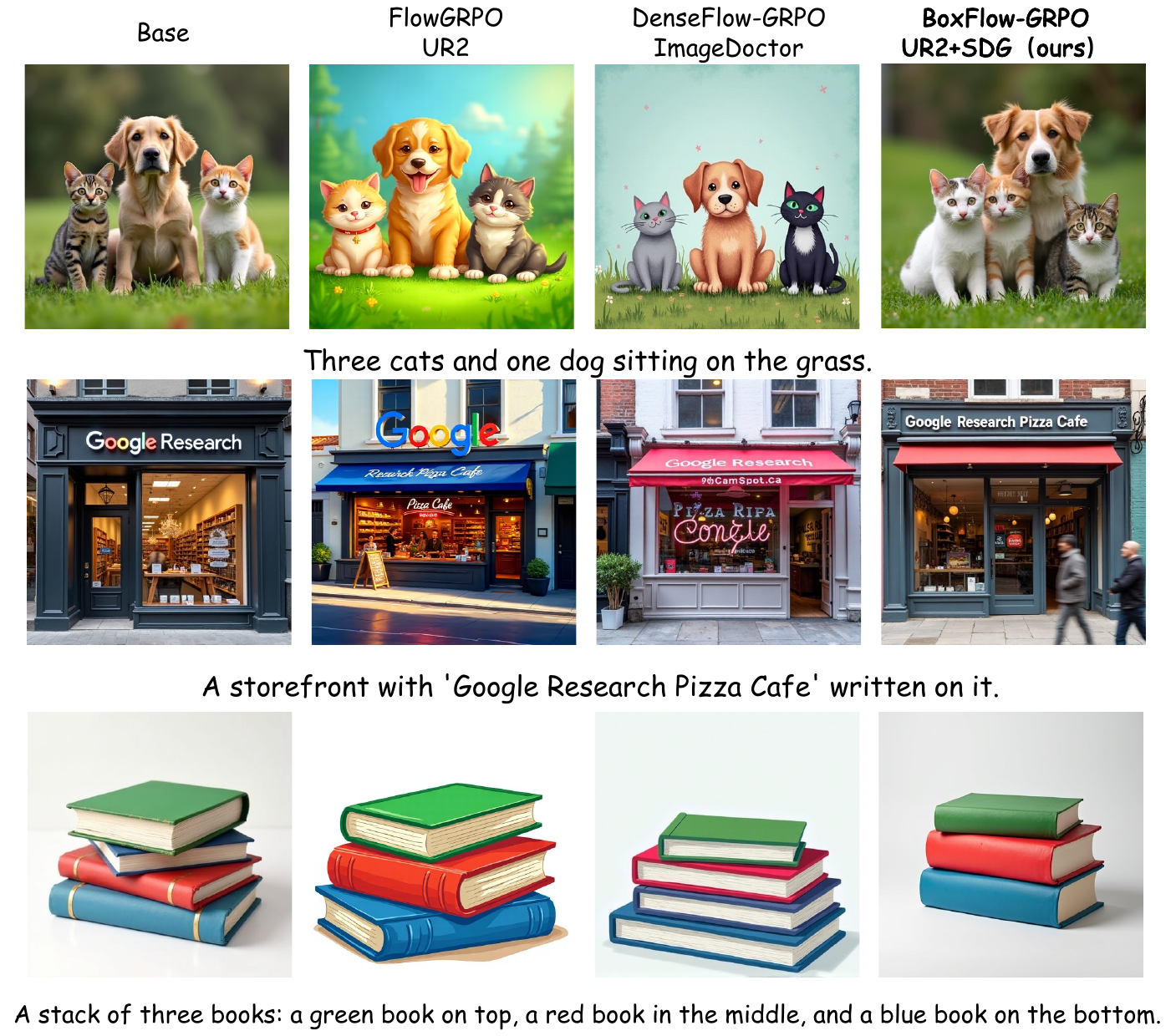}
\vspace{-6pt}
\caption{Qualitative comparison of BoxFlow-GRPO.}
\vspace{-15pt}
\label{fig:denseflow_vis}
\end{minipage}
\hfill
\begin{minipage}[t]{0.49\textwidth}
\centering
\includegraphics[width=0.82\textwidth]{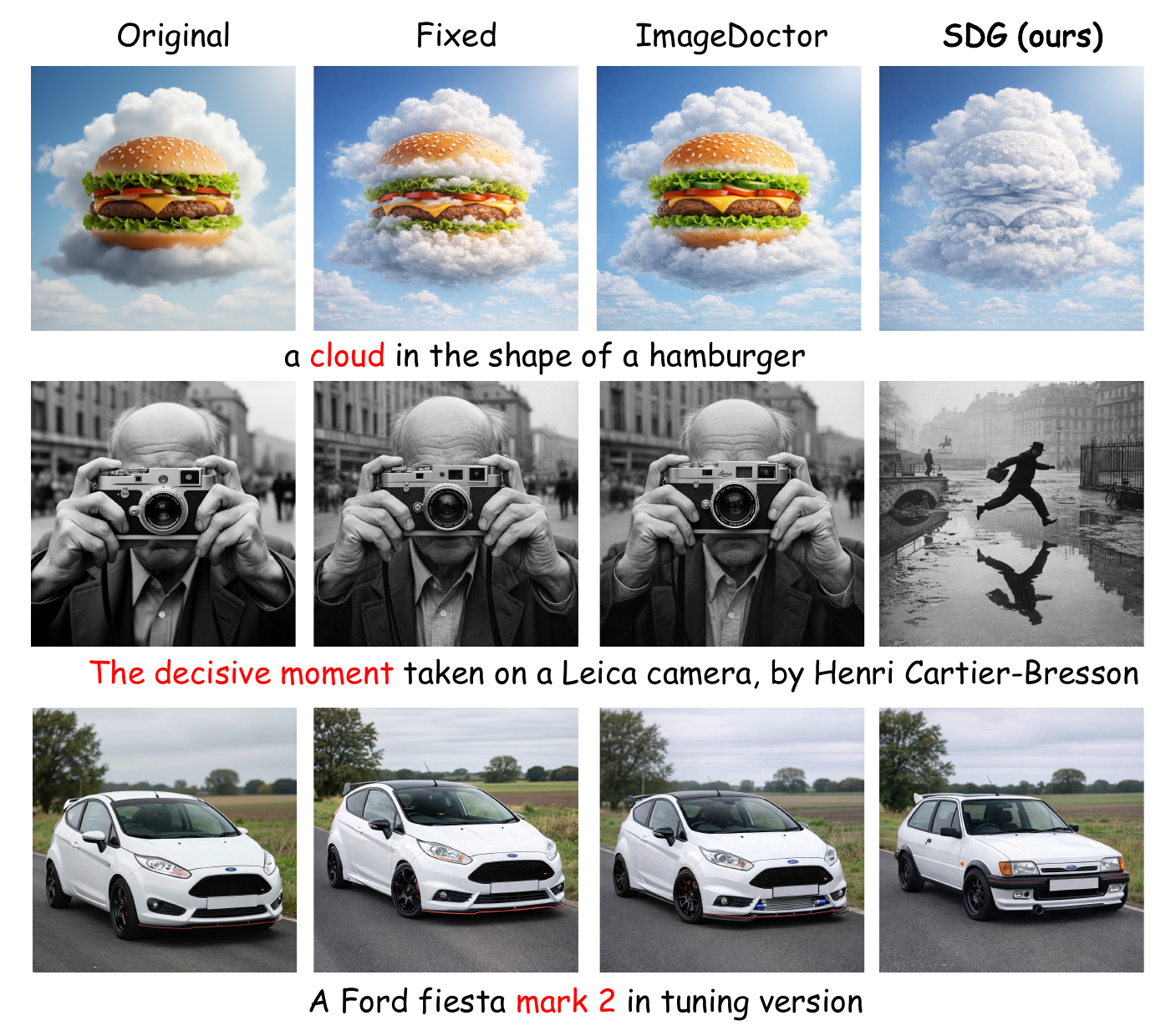}
\vspace{-6pt}
\caption{Qualitative comparison of defect-guided image refinement via GPT-Image-1.5.}
\label{fig:reflection}
\end{minipage}
\vspace{-8pt}
\end{figure}

\textbf{Results.}
Table~\ref{tab:ur2_box_results} shows that baseline RL variants can increase reward-model scores while drifting toward illustration- or anime-like outputs, with lower $P(\mathrm{real})$ than Base. BoxFlow-GRPO achieves the best average relative change ($+2.4\%$) and highest $P(\mathrm{real})$ (0.228), while maintaining competitive preference and quality scores. Figure~\ref{fig:denseflow_vis} illustrates reduced artifacts and misalignment with preserved realism. On Qwen-Image, after 1,200 training steps, BoxFlow-GRPO again achieves the best average relative improvement ($+1.1\%$; Appendix~\ref{app:qwen_image_results}).

\begin{table}[t]
\centering
\caption{Downstream performance of BoxFlow-GRPO and baselines. Parentheses show changes relative to Base (\textcolor{green!60!black}{green}: improvement; \textcolor{red!80!black}{red}: regression); Avg averages these changes. Only BoxFlow-GRPO improves all five dimensions.}
\label{tab:ur2_box_results}
\footnotesize
\setlength{\tabcolsep}{3pt}
\renewcommand{\arraystretch}{1.2}
\resizebox{\linewidth}{!}{%
\begin{tabular}{ll|cccccc}
\toprule
Method & Reward & PickScore & CLIPScore & HPSv3 & DeQA & $P(\mathrm{real})$ & Avg \\
\midrule
Base & -- & 22.84 & 0.912 & 11.75 & 0.876 & 0.211 & -- \\
\midrule
\multirow{2}{*}{Flow-GRPO}
  & UR2 & 22.78\,\textcolor{red!80!black}{\tiny(-0.3\%)} & 0.908\,\textcolor{red!80!black}{\tiny(-0.4\%)} & 11.27\,\textcolor{red!80!black}{\tiny(-4.1\%)} & \textbf{0.882}\,\textcolor{green!60!black}{\tiny(+0.7\%)} & 0.149\,\textcolor{red!80!black}{\tiny(-29.4\%)} & \textcolor{red!80!black}{-6.7\%} \\
  & ImageDoctor & \textbf{23.17}\,\textcolor{green!60!black}{\tiny(+1.4\%)} & \underline{0.937}\,\textcolor{green!60!black}{\tiny(+2.7\%)} & \textbf{12.23}\,\textcolor{green!60!black}{\tiny(+4.1\%)} & 0.857\,\textcolor{red!80!black}{\tiny(-2.2\%)} & \underline{0.201}\,\textcolor{red!80!black}{\tiny(-4.7\%)} & \underline{\textcolor{green!60!black}{+0.3\%}} \\
\midrule
DenseFlow-GRPO & ImageDoctor & 22.88\,\textcolor{green!60!black}{\tiny(+0.2\%)} & \textbf{0.945}\,\textcolor{green!60!black}{\tiny(+3.6\%)} & 11.57\,\textcolor{red!80!black}{\tiny(-1.5\%)} & 0.864\,\textcolor{red!80!black}{\tiny(-1.4\%)} & 0.199\,\textcolor{red!80!black}{\tiny(-5.7\%)} & \textcolor{red!80!black}{-1.0\%} \\
\midrule
\textbf{BoxFlow-GRPO} (ours) & UR2+SDG & \underline{22.91}\,\textcolor{green!60!black}{\tiny(+0.3\%)} & 0.915\,\textcolor{green!60!black}{\tiny(+0.3\%)} & \underline{12.14}\,\textcolor{green!60!black}{\tiny(+3.4\%)} & \underline{0.877}\,\textcolor{green!60!black}{\tiny(+0.1\%)} & \textbf{0.228}\,\textcolor{green!60!black}{\tiny(+8.1\%)} & \textbf{\textcolor{green!60!black}{+2.4\%}} \\
\bottomrule
\end{tabular}%
}
\vspace{-10pt}
\end{table}

\subsubsection{Defect-Guided Image Refinement}
\label{sec:reflection_results}

\textbf{Setup.}
We use the GRPO checkpoint of the SDG detector to diagnose defects, then provide GPT-Image-1.5 with a box overlay and structured text feedback. We compare it with \textbf{Fixed} caption-only editing and \textbf{ImageDoctor} heatmap-based feedback, using the same editor for all methods. Two annotators independently and blindly compare paired outputs on 873 valid samples retained after GPT-Image-1.5 filtering, assigning Good/Same/Bad labels following HunyuanImage 3.0~\citep{cao2026hunyuanimage30technicalreport}, where Good means SDG is preferred.

\begin{wraptable}{r}{0.48\textwidth}
\vspace{-18pt}
\centering
\caption{GSB rates between SDG and baseline methods over 873 valid samples.}
\label{tab:gsb_all_samples}
\footnotesize
\setlength{\tabcolsep}{2pt}
\renewcommand{\arraystretch}{1.15}
\resizebox{\linewidth}{!}{%
\begin{tabular}{lccc}
\toprule
Comparison & Good (\%) & Same (\%) & Bad (\%) \\
\midrule
\textbf{SDG} vs ImageDoctor & 11.00 & 85.11 & 3.90 \\
\textbf{SDG} vs Fixed & 10.31 & 86.94 & 2.75 \\
\bottomrule
\end{tabular}%
}
\vspace{-12pt}
\end{wraptable}
\textbf{Results.}
Table~\ref{tab:gsb_all_samples} shows higher Good than Bad rates against ImageDoctor (11.00\% vs.\ 3.90\%) and Fixed (10.31\% vs.\ 2.75\%). The high Same rate (85.1--86.9\%) reflects the strong editor and high-quality inputs. Re-diagnosing paired images with the same frozen SDG detector shows that SDG-guided refinement reduces detected boxes by 37.55\% and total importance by 42.37\%, the largest reductions among the three methods (Appendix~\ref{app:refinement_rediagnosis}). This complements the human preference results with detector-level evidence of effective refinement. Qualitative examples appear in Figure~\ref{fig:reflection} and Appendix~\ref{app:reflection_extended}.




\section{Conclusion}
\label{sec:conclusion}

We presented \textbf{Structured Defect Grounding (SDG)}, an instance-level formulation for dense text-to-image (T2I) feedback that models each defect as a \emph{(location, type, reason, importance)} tuple and casts diagnosis as variable-cardinality structured set prediction. This formulation supports \textbf{SDG-30K}, a 30K-image box-grounded dataset across four modern T2I generators, and \textbf{SDG-Eval} for structured defect-set evaluation. It also enables a diagnosis-to-alignment pipeline in which a VLM-based detector predicts defect sets and \textbf{BoxFlow-GRPO} converts them into dense spatial rewards for diffusion alignment. Experiments demonstrate that SDG outperforms leading proprietary VLMs in zero-shot defect grounding, improves T2I alignment, and supports actionable defect-guided image refinement, making it a practical interface for evaluating and improving modern generative models.

\begin{ack}
This work was supported by the SSTIC Grants (KJZD20230923115106012, KJZD20230923114916032, GJHZ20240218113604008).
\end{ack}

\bibliographystyle{plainnat}
\bibliography{references}

\newpage
\appendix
\section{Annotation Guidelines}
\label{app:annotation-guideline}

This appendix describes the annotation protocol used to construct the SDG-30K dataset. The annotation task was conducted as data curation for generated images rather than as a user study. Annotators were compensated above the local minimum wage in the data-collection location.

\subsection{Interface and Workflow}

The annotation interface consists of two synchronized panels: a raw image canvas on which annotators draw the final bounding boxes and a metadata panel showing the generation prompt together with annotation controls.

Each image is processed in two passes. In the \emph{initial annotation pass}, annotators examine the prompt-image pair, draw defect boxes from scratch, assign top-level labels, and write concise descriptions for each confirmed defect. In the \emph{global scan pass}, annotators re-examine the full image to identify missed defects, adjust box boundaries, and improve annotation completeness. No machine-generated candidate boxes or defect descriptions are shown during manual annotation.

\subsection{Defect-Type Taxonomy}

We define two top-level defect categories, each with fine-grained subtypes.

\textbf{Misalignment} captures inconsistencies between the generation prompt and image content. Subtypes include missing objects, extra objects, attribute mismatches (color, count, material), spatial-relation errors, action mismatches, and style mismatches. For example, a prompt requesting ``two cats on a red sofa'' paired with an image showing three cats constitutes a count-based misalignment.

\textbf{Artifact} captures visual plausibility defects independent of the prompt. Subtypes include anatomical distortions (e.g., malformed hands), geometric deformations, texture abnormalities, edge and contour defects, text garbling, and lighting inconsistencies. For example, fused fingers or impossible joint angles constitute anatomical artifacts regardless of the prompt content.

The taxonomy explicitly separates semantic mismatch from visual corruption: green eyes are a misalignment when the prompt requests blue eyes, while geometrically distorted eyes are artifacts. Intentional artistic stylization (e.g., cubist deformation) is not labeled as a defect; only non-intentional structural or perceptual inconsistencies qualify.

\subsection{Boxing and Description Principles}

Although bounding boxes are coarser than pixel masks, they provide a practical trade-off between spatial specificity, annotation efficiency, and compatibility with VLM outputs.
Bounding boxes should tightly cover the defective location while excluding unnecessary background. Each box corresponds to one concrete defect: when a large defect location contains multiple independent issues, annotators split them into separate boxes; when multiple small detections reflect a single coherent failure mode, they may be merged.

For each box, annotators provide a concise reason statement (target length $\le$30 Chinese characters) that specifically describes the defect. Vague comments such as ``looks wrong'' are prohibited; descriptions must reference concrete visual evidence (e.g., ``left hand has six fingers'' or ``hat mentioned in prompt is absent'').

\subsection{Clean Images}

The annotation platform required at least one bounding box per submission. For images without defects, annotators drew a placeholder box and assigned an explicit \emph{no-problem} tag. During post-processing, these samples are converted to empty defect sets, preserving them as valid negative examples for training and evaluation.

\subsection{Prompt Interpretation Protocol}

To ensure the correctness of prompt interpretation, annotators were instructed to consult search engines whenever a prompt involved unfamiliar named entities, cultural references, artistic styles, products, landmarks, or other domain-specific concepts. External search was used only to clarify the semantic background of the prompt, rather than to provide defect labels directly. This protocol reduces annotation noise caused by incomplete prior knowledge and improves the consistency of prompt-conditioned \textit{misalignment} annotation.

\subsection{Importance Scoring Rubric}

Each defect instance receives an importance score (integer from 1 to 100) reflecting how much it affects overall image quality and caption faithfulness, assessed via a rubric-guided Gemini protocol. The four criteria are considered in order of priority:
\begin{enumerate}
\item \emph{Visual prominence} --- how easily a typical viewer can spot the defect at normal viewing distance.
\item \emph{Semantic impact} --- whether the defect changes the meaning, identity, or key content relative to the prompt.
\item \emph{Area coverage} --- larger defects affecting more of the image score higher.
\item \emph{Location} --- defects on the main subject or focal area score higher than those in the background.
\end{enumerate}
Scores are grouped into five tiers: Critical (90--100), Major (70--89), Moderate (40--69), Minor (15--39), and Negligible (1--14). Importance annotations are obtained by prompting Gemini with the image, the original prompt, and the human-annotated defect set, and are used both as SFT supervision and as a GRPO reward signal.

\section{Prompt Templates}
\label{app:prompts}

This appendix provides the full prompt templates used in SDG. The same user prompt (Appendix~\ref{app:sft_prompt}) is used for SFT training, GRPO training, and inference. The system prompt uses the default Qwen chat template (``You are a helpful assistant.'').

\subsection{SFT / GRPO / Inference User Prompt}
\label{app:sft_prompt}

\begin{tcolorbox}[breakable, colback=blue!3!white, colframe=blue!50!black, title=SDG User Prompt, left=3pt, right=3pt, top=2pt, bottom=2pt]
\small\texttt{You are an AI image quality evaluator. You will be given \textbf{one image} to analyze.}

\smallskip
\small\phead{\#\#\# Definitions}

\smallskip
\small\texttt{\textbf{Misalignment}: Areas where the image content does NOT match the text caption, including:}\\
\small\texttt{- Missing objects: Objects mentioned in caption but not present in image}\\
\small\texttt{- Extra objects: Objects present in image but not mentioned in caption}\\
\small\texttt{- Wrong attributes: Incorrect color, size, material, count, or other properties}\\
\small\texttt{- Wrong spatial relationships: Incorrect positions, orientations, or arrangements}

\smallskip
\small\texttt{\textbf{Artifact}: Visual defects in the generated image, including:}\\
\small\texttt{- Distorted anatomy: Malformed hands, extra/missing limbs, wrong number of fingers}\\
\small\texttt{- Duplicated/missing parts: Repeated or absent body parts, objects}\\
\small\texttt{- Warped geometry: Perspective errors, impossible shapes}\\
\small\texttt{- Texture issues: Melted, smeared, or overly smooth textures}\\
\small\texttt{- Unnatural edges: Jagged, broken, or blurry boundaries}\\
\small\texttt{- Garbled text: Unreadable or malformed text/letters}\\
\small\texttt{- Lighting inconsistencies: Wrong shadows, reflections, or light sources}

\smallskip
\small\texttt{Text Caption: \{caption\}}

\smallskip
\small\texttt{\textbf{Goal}: Produce a detailed analysis of the image quality and output bounding boxes with severity scores for all detected issues.}

\smallskip
\small\phead{\#\#\# Strict Output Rules}\\
\small\texttt{Output \textbf{TWO blocks in this exact order}:}\\
\small\texttt{1) \xmltag{<think>} - Your detailed analysis}\\
\small\texttt{2) \xmltag{<answer>} - JSON list of bounding boxes}

\smallskip
\small\phead{\#\#\# Think Format (STRICT)}\\
\small\texttt{\xmltag{<think>}}\\
\small\phead{\#\#\# Step 1: Caption Understanding}\\
\small\texttt{- Briefly summarize what the caption requires (subject, key attributes, actions, setting, style/composition if mentioned).}

\smallskip
\small\phead{\#\#\# Step 2: Visual Analysis \& Defect Spotting (Issue Summary)}\\
\small\texttt{- Describe the quality issues you observe in the image.}\\
\small\texttt{- You MAY merge similar issues into a single bullet.}\\
\small\texttt{- Each bullet MUST include:}\\
\small\texttt{\quad(a) the issue category (artifact or misalignment, or both if needed)}\\
\small\texttt{\quad(b) what is affected (object/part)}\\
\small\texttt{\quad(c) concrete visual evidence (what specifically looks wrong/missing/mismatched)}\\
\small\texttt{- Do NOT mention numeric coordinates.}

\smallskip
\small\phead{\#\#\# Step 3: Localization (Box-by-Box Grounding)}\\
\small\texttt{- Provide a detailed, precise localization statement for EACH defect instance (one line per box).}\\
\small\texttt{- Do NOT mention numeric coordinates.}\\
\small\texttt{- Each localization line MUST include all of the following information in natural language:}\\
\small\texttt{\quad1) Anchor: the exact object/part involved}\\
\small\texttt{\quad2) Position: image-based cues (image-left/right, upper/lower, center, near the border)}\\
\small\texttt{\quad3) Interaction cue (when applicable): holding/\allowbreak touching/\allowbreak overlapping/\allowbreak merging/\allowbreak etc.}\\
\small\texttt{\quad4) Scale description: tiny localized detail, part-sized area, large area, or whole image}\\
\small\texttt{\quad5) Shape/orientation: compact, elongated, runs along a boundary, wraps around an object}\\
\small\texttt{- Do NOT invent new defects; each line must correspond to exactly one defect instance.}\\
\small\texttt{\xmltag{</think>}}

\smallskip
\small\phead{\#\#\# Answer Format (for \xmltag{<answer>})}\\
\small\texttt{{[}\{``box\_2d'': {[}x0, y0, x1, y1{]}, ``label'': ``misalignment''|``artifact'', ``description'': ``brief description of the issue'', ``importance'': N\}{]}}

\smallskip
\small\texttt{Bounding box coordinates are in normalized 0--1000 space: {[}x0, y0, x1, y1{]}.}\\
\small\texttt{If there are no issues, output an empty list.}

\smallskip
\small\phead{\#\#\# Importance Scoring}\\
\small\texttt{For EACH box, assign an integer importance score from 1 to 100:}\\
\small\texttt{- 90--100: Critical --- immediately obvious, ruins the image.}\\
\small\texttt{- 70--89: Major --- clearly visible at normal viewing distance.}\\
\small\texttt{- 40--69: Moderate --- noticeable on closer inspection.}\\
\small\texttt{- 15--39: Minor --- only visible on careful examination.}\\
\small\texttt{- 1--14: Negligible --- barely perceptible.}

\smallskip
\small\phead{\#\#\# Description Style Guide}\\
\small\texttt{- 18 to 45 words per description.}\\
\small\texttt{- Must mention the affected object/part.}\\
\small\texttt{- Must include at least one concrete evidence phrase.}\\
\small\texttt{- Do NOT use vague words like ``weird/strange/bad''.}\\
\small\texttt{- Do NOT include numeric coordinates in description.}

\smallskip
\small\texttt{Now analyze the image and produce your output:}
\end{tcolorbox}

\subsection{Gemini Distillation Prompt}
\label{app:gemini_prompt}

During data preparation, we use Gemini 3 Pro to translate Chinese defect descriptions into English, expand them with richer detail, and assign importance scores. The model receives the image, caption, and human-annotated defect boxes as hidden context. The full prompt is shown below.

\begin{tcolorbox}[breakable, colback=orange!3!white, colframe=orange!60!black, title=Gemini Distillation Prompt, left=3pt, right=3pt, top=2pt, bottom=2pt]
\small\texttt{You are an AI image quality evaluator. You will be shown an image and asked to identify quality issues.}

\smallskip
\small\texttt{Hidden context (DO NOT reveal):}\\
\small\texttt{The human annotations below are Ground Truth. You MUST keep the same number of boxes, the same coordinates, and the same labels in the final JSON. You must behave as if you discovered the issues independently from the image and caption.}

\smallskip
\small\texttt{Ground Truth annotations (DO NOT modify coordinates or labels; you may rewrite/expand description in English):}\\
\small\texttt{- Artifact boxes: \{artifact\_bbox\_text\}}\\
\small\texttt{- Misalignment boxes: \{misalignment\_bbox\_text\}}

\smallskip
\small\texttt{Text Caption: \{caption\}}

\smallskip
\small\texttt{====== GLOBAL CONSTRAINTS (MUST FOLLOW) ======}\\
\small\texttt{1) NEVER mention annotations, boxes, ground truth, translation, or any external hints.}\\
\small\texttt{2) Do NOT invent extra defects or remove any defect. The final JSON must contain exactly the same set of boxes/labels as provided.}\\
\small\texttt{3) IMPORTANT: box\_2d uses RELATIVE coordinates on a 0--1000 scale.}\\
\small\texttt{\quad- Format: {[}x\_min, y\_min, x\_max, y\_max{]} (xyxy)}\\
\small\texttt{\quad- You MUST output these coordinates exactly as given in the final JSON.}\\
\small\texttt{4) Do NOT write numeric coordinates in the \xmltag{<think>} block.}\\
\small\texttt{5) Output MUST contain exactly TWO blocks in this order:}\\
\small\texttt{\quad(1) \xmltag{<think>} ... \xmltag{</think>}\quad(2) \xmltag{<answer>} ... \xmltag{</answer>}}\\
\small\texttt{6) \xmltag{<answer>} MUST be a JSON list in xyxy format:}\\
\small\texttt{\quad{[}\{``box\_2d'':{[}x0,y0,x1,y1{]},``label'':``artifact''|``misalignment'',}\\
\small\texttt{\quad\quad``description'':``...'',``importance'':N\}, ...{]}}

\smallskip
\small\texttt{====== IMPORTANCE SCORING (FIELD: ``importance'') ======}\\
\small\texttt{For EACH box, assign an integer importance score from 1 to 100:}\\
\small\texttt{- 90--100: Critical --- immediately obvious, ruins the image.}\\
\small\texttt{- 70--89: Major --- clearly visible at normal viewing distance.}\\
\small\texttt{- 40--69: Moderate --- noticeable on closer inspection.}\\
\small\texttt{- 15--39: Minor --- only visible on careful examination.}\\
\small\texttt{- 1--14: Negligible --- barely perceptible.}

\smallskip
\small\texttt{Scoring criteria (in order of priority):}\\
\small\texttt{a) Visual prominence --- How easily can a typical viewer spot this defect at normal viewing distance?}\\
\small\texttt{b) Semantic impact --- Does this defect change the meaning, identity, or key content of the image relative to the caption?}\\
\small\texttt{c) Area coverage --- Larger defects affecting more of the image score higher.}\\
\small\texttt{d) Location --- Defects on the main subject or focal area score higher than those in background/periphery.}

\smallskip
\small\texttt{====== THINKING FORMAT (STRICT) ======}\\
\small\texttt{\xmltag{<think>}}\\
\small\phead{\#\#\# Step 1: Caption Understanding}\\
\small\phead{\#\#\# Step 2: Visual Analysis \& Defect Spotting (Issue Summary)}\\
\small\phead{\#\#\# Step 3: Localization (Box-by-Box Grounding)}\\
\small\texttt{\xmltag{</think>}}

\smallskip
\small\texttt{====== DESCRIPTION STYLE GUIDE (FOR \xmltag{<answer>}) ======}\\
\small\texttt{For EACH box in the final JSON, rewrite description into English with richer detail:}\\
\small\texttt{- 18 to 45 words per description.}\\
\small\texttt{- Must mention the affected object/part.}\\
\small\texttt{- Must include at least one concrete evidence phrase.}\\
\small\texttt{- Do NOT use vague words like ``weird/strange/bad''.}\\
\small\texttt{- Do NOT mention translation or ground truth.}

\smallskip
\small\texttt{====== FEW-SHOT EXAMPLES ======}

\smallskip
\small\texttt{{[}EXAMPLE A: NON-EMPTY GT{]}}\\
\small\texttt{Caption: a rock musician sticking tongue out holding a sign that says hail Satan}

\smallskip
\small\texttt{\xmltag{<think>}}\\
\small\phead{\#\#\# Step 1: Caption Understanding}\\
\small\texttt{- The caption asks for a rock musician sticking their tongue out while holding a sign reading ``hail Satan''.}\\
\small\phead{\#\#\# Step 2: Visual Analysis \& Defect Spotting}\\
\small\texttt{- Misalignment: The person does not convincingly match the specified celebrity identity.}\\
\small\texttt{- Misalignment: The sign text is not rendered as the requested phrase.}\\
\small\texttt{- Artifact: Ring edges blend into skin and local finger texture appears unnaturally bumpy.}\\
\small\phead{\#\#\# Step 3: Localization}\\
\small\texttt{- Box 1 (artifact): Ring--skin contact on image-right hand, tiny localized detail.}\\
\small\texttt{- Box 2 (artifact): Mid-finger ring edge, compact boundary location.}\\
\small\texttt{- Box 3 (artifact): Below rings on image-right fingers, tiny skin patch.}\\
\small\texttt{- Box 4 (artifact): Image-left hand ring-to-finger junction, tiny localized.}\\
\small\texttt{- Box 5 (misalignment): Upper-central face location, part-sized.}\\
\small\texttt{- Box 6 (misalignment): Central sign surface with lettering, part-sized.}\\
\small\texttt{\xmltag{</think>}}\\
\small\texttt{\xmltag{<answer>}}\\
\small\texttt{{[}\{``box\_2d'':{[}777,604,796,627{]}, ``label'':``artifact'',}\\
\small\texttt{\quad``description'':``Ring appears fused into finger with smeared metal-to-skin boundary...'',}\\
\small\texttt{\quad``importance'':45\}, ...{]}}\\
\small\texttt{\xmltag{</answer>}}

\smallskip
\small\texttt{{[}EXAMPLE B: EMPTY GT{]}}\\
\small\texttt{Caption: a girl with pink curly hair wearing a witch hat}\\
\small\texttt{\xmltag{<think>} ... No defects identified. \xmltag{</think>}}\\
\small\texttt{\xmltag{<answer>} {[}{]} \xmltag{</answer>}}

\smallskip
\small\texttt{Now analyze the image and respond with exactly TWO blocks (\xmltag{<think>} then \xmltag{<answer>}).}
\end{tcolorbox}

\section{Evaluation Metric Details}
\label{app:eval_metrics}

Let $\mathcal{T}=\{\textit{artifact},\textit{misalignment}\}$ and let $\mathcal{D}$ be the evaluation set. For image $d$, denote the ground-truth and predicted defect sets as
$G(d)=\{(b_i,t_i,r_i,s_i)\}_{i=1}^{N_d}$ and
$P(d)=\{(\hat b_j,\hat t_j,\hat r_j,\hat s_j)\}_{j=1}^{M_d}$.
For type $t$, let $G_t(d)$ and $P_t(d)$ be the subsets whose type is $t$.

\paragraph{Image-level metrics.}
We define binary indicators $y_{t,d}=\mathbf{1}[|G_t(d)|>0]$ and $\hat y_{t,d}=\mathbf{1}[|P_t(d)|>0]$.
DetTypeF1 is the type-specific image-level F1:
\begin{equation}
Pr_t=\frac{\sum_{d\in\mathcal{D}} y_{t,d}\hat y_{t,d}}{\sum_{d\in\mathcal{D}}\hat y_{t,d}},\quad
Re_t=\frac{\sum_{d\in\mathcal{D}} y_{t,d}\hat y_{t,d}}{\sum_{d\in\mathcal{D}} y_{t,d}},\quad
\mathrm{DetTypeF1}_t=\frac{2Pr_tRe_t}{Pr_t+Re_t}.
\end{equation}
Clean-image accuracy is the true-negative rate on $\mathcal{D}^-_t=\{d\in\mathcal{D}\mid y_{t,d}=0\}$:
\begin{equation}
\mathrm{ClnAcc}_t=\frac{1}{|\mathcal{D}^-_t|}\sum_{d\in\mathcal{D}^-_t}\mathbf{1}[\hat y_{t,d}=0].
\end{equation}

\paragraph{Defect-level metrics.}
For each $d$, we compute a class-aware Hungarian matching~\citep{kuhn1955hungarian} between $G(d)$ and $P(d)$ under the constraint $t_g=t_p$.
Let $\mathcal{M}_{\tau,t}$ be the union of matched pairs of type $t$ whose IoU is at least $\tau$.
For $\tau\in\{0.1,0.5\}$, localization precision, recall, and F1 are
\begin{equation}
\mathrm{BoxPr}_t=\frac{|\mathcal{M}_{\tau,t}|}{\sum_{d\in\mathcal{D}}|P_t(d)|},\quad
\mathrm{BoxRe}_t=\frac{|\mathcal{M}_{\tau,t}|}{\sum_{d\in\mathcal{D}}|G_t(d)|},\quad
\mathrm{BoxF1}_t=\frac{2|\mathcal{M}_{\tau,t}|}{\sum_{d\in\mathcal{D}}(|G_t(d)|+|P_t(d)|)}.
\end{equation}
For each valid matched pair $(g,p)\in\mathcal{M}_{\tau,t}$, DescCos is the mean cosine similarity between Qwen3-Embedding-0.6B embeddings of $r_g$ and $\hat r_p$, and ImpAcc is normalized absolute-error accuracy:
\begin{equation}
\mathrm{DescCos}_t=\frac{1}{|\mathcal{M}_{\tau,t}|}\sum_{(g,p)\in\mathcal{M}_{\tau,t}}\langle r_g,\hat r_p\rangle,\quad
\mathrm{ImpAcc}_t=\frac{1}{|\mathcal{M}_{\tau,t}|}\sum_{(g,p)\in\mathcal{M}_{\tau,t}}\left(1-\frac{|s_g-\hat s_p|}{100}\right).
\end{equation}

\section{Experimental Details}
\label{app:experimental_details}

This appendix provides the complete reward computation and training details for the GRPO stage described in Section~\ref{sec:training}.

\subsection{Reward Formulation}
\label{app:reward_formulation}

\paragraph{Composite reward.}
The composite reward is gated by a format check:
\begin{equation}
R = \begin{cases}
0.6\, R_{\mathrm{diou}} + 0.25\, R_{\mathrm{desc}} + 0.15\, R_{\mathrm{imp}}, & \text{if format valid,} \\
-1, & \text{otherwise.}
\end{cases}
\end{equation}
The format gate verifies that the response contains valid \texttt{<think>} tags, a parseable JSON defect list in \texttt{<answer>}, and properly ordered \texttt{xyxy} coordinates.

\paragraph{Grounding accuracy ($R_{\mathrm{diou}}$).}
Predicted and ground-truth boxes are matched using the Hungarian algorithm with DIoU as the cost metric, yielding optimal one-to-one assignments. For each matched pair $(i,j)\in\mathcal{M}$, the spatial reward is the DIoU score of that pair.

Edge cases are handled as follows:
\begin{itemize}
\item \textbf{Correct rejection} (both sets empty): $R_{\mathrm{diou}}=0.3$.
\item \textbf{Miss} (ground-truth defects exist but no predictions): $R_{\mathrm{diou}}=-0.8$.
\item \textbf{False alarm} (predictions on a clean image): $R_{\mathrm{diou}}=-0.3$.
\item \textbf{Unmatched boxes}: each receives a penalty of $-0.5$.
\end{itemize}
The final score is normalized by $\max(|G|, |P|, 1)$ and clipped to $[-1, 1]$.

\paragraph{Description consistency ($R_{\mathrm{desc}}$).}
For each matched pair $(i,j) \in \mathcal{M}$, we compute the cosine similarity between the predicted and ground-truth descriptions using Qwen3-Embedding-0.6B. The raw similarity is linearly transformed to $[0,1]$ via:
\begin{equation}
\hat{s}_{ij} = \mathrm{clip}\!\left(\frac{\mathrm{sim}(r_i, \hat{r}_j) - 0.5}{0.4}, \;0, \;1\right).
\end{equation}
The description reward is the sum of transformed similarities over matched pairs, divided by $\max(|G|, |P|, 1)$. Unmatched boxes contribute zero, implicitly penalizing over- or under-prediction.

\paragraph{Importance estimation ($R_{\mathrm{imp}}$).}
For each matched pair $(i,j) \in \mathcal{M}$, the importance reward is:
\begin{equation}
r_{\mathrm{imp}}^{(i,j)} = \mathrm{clip}\!\left(1 - \frac{|\hat{s}_j - s_i|}{50}, \;0, \;1\right),
\end{equation}
where $s_i$ and $\hat{s}_j$ are the ground-truth and predicted importance scores. This provides a continuous reward that decreases linearly with absolute error, reaching zero when the error exceeds 50 points.

\subsection{GRPO Objective}

The full GRPO objective optimizes the policy using clipped importance ratios and KL regularization:
\begin{equation}
\mathcal{L}_{\mathrm{GRPO}} =
-\mathbb{E}\left[
\min\left(
\rho_s A_s,\;
\mathrm{clip}(\rho_s, 1-\epsilon, 1+\epsilon) A_s
\right)
\right]
+ \beta\, \mathrm{KL}(\pi_\theta \,\|\, \pi_{\mathrm{ref}}),
\end{equation}
where
\begin{equation}
\rho_s = \frac{\pi_\theta(y_s \mid I,T)}{\pi_{\mathrm{old}}(y_s \mid I,T)}
\end{equation}
is the importance ratio for sampled response $y_s$, $\epsilon$ is the clipping range, and $\beta{=}0.01$ is the KL regularization coefficient.

\subsection{SFT Hyperparameters}

All generated images are resized so that the longer side is at most 1024 pixels. We use DeepSpeed ZeRO-2 with bfloat16 mixed precision on 2~nodes $\times$ 8 GPUs (16 A100-80G GPUs total). The learning rate is $3 \times 10^{-5}$ with a cosine schedule and 5\% warmup. We train for 1 epoch with per-device batch size 1 and gradient accumulation steps 1, yielding an effective batch size of 16; the $3{\times}$ pre-baked jitter augmentation effectively exposes the model to three passes over each example within this single epoch. The maximum sequence length is 5,100 tokens, and the vision encoder is frozen throughout.

For coordinate jitter augmentation, each coordinate receives an independent random offset sampled uniformly from $[-10, +10]$ in the $[0,1000]$ normalized space, with clamping to ensure valid box constraints. We independently sample the offsets when preparing the three augmented copies, providing coordinate variation across repeated presentations of each example.

\subsection{GRPO Hyperparameters}

We use 2 nodes $\times$ 8 GPUs (16 GPUs total) with DeepSpeed ZeRO-2 and a learning rate of $5 \times 10^{-6}$. The KL coefficient is $\beta{=}0.01$. For each prompt, $S{=}8$ candidate responses are sampled via colocated vLLM rollout with temperature 1.0 and top-$p$ 0.85 (max completion length 4,096 tokens). We train for 2 epochs with per-device batch size 4.

\subsection{Compute Resources}
\label{app:compute_resources}

All detector training experiments are run on A100-80G GPUs. A single SDG detector SFT run takes approximately 2 hours on $16{\times}$A100-80G GPUs, while a single detector GRPO run takes approximately 36 hours on $16{\times}$A100-80G GPUs. For diffusion alignment, one BoxFlow-GRPO run takes approximately 24 hours on $16{\times}$A100 GPUs. The defect-guided image refinement experiments use GPT-Image-1.5 through an external API; their wall-clock time depends primarily on API throughput and the allowed request concurrency rather than local GPU compute. The estimated wall-clock time for the reported experiments is about 7 days in total. The full research project required additional compute beyond this estimate because it included preliminary studies and failed experimental runs that are not reported in the paper.

\subsection{Data, Code, and Model Availability}
\label{app:data_code_availability}

The \href{https://github.com/nianbai006/SDG}{public code repository} provides source code and reproduction instructions. The \href{https://huggingface.co/collections/P1n3/sdg}{SDG collection} provides SDG-30K, the SFT detector, the merged GRPO detector, and the BoxFlow-GRPO FLUX adapter. The GRPO detector is a full merged checkpoint and should be loaded directly rather than as a LoRA adapter.

\subsection{Existing Assets and Licenses}
\label{app:asset_licenses}

We use existing assets only for research dataset construction, training, evaluation, or API-based editing, and cite their original sources throughout the paper. Pick-a-Pic prompts are used under the MIT License. Qwen3-VL-4B-Instruct and Qwen3-Embedding-0.6B are released under Apache-2.0. The T2I generators used for SDG-30K follow their respective public licenses or terms: FLUX.2-dev is governed by the FLUX Non-Commercial License, Z-Image-Turbo and LongCat-Image are released under Apache-2.0, and SANA-1.5 is released under NSCL v2-custom / NVIDIA License. Gemini 3 Pro, GPT-5.4, and GPT-Image-1.5 are accessed only through their official API terms. Public baselines and evaluation resources, including ImageDoctor, Flow-GRPO, UnifiedReward-2.0, PickScore, CLIPScore, HPSv3, DeQA, Forensic-Chat, DrawBench, and RichHF-18K, are cited and used according to their public releases or provider terms. We do not redistribute third-party weights, prompts, images, or code except where allowed by their licenses; released SDG assets will include attribution, license notices, intended-use documentation, and pointers to the original sources.

\subsection{LLM Usage Declaration}
\label{app:llm_usage}

LLMs and VLMs are used as core components of this work. Gemini 3 Pro is used during data preparation for description expansion, reasoning-trace distillation, and importance scoring; Qwen3-VL-4B-Instruct is fine-tuned as the SDG detector; Qwen3-Embedding-0.6B is used for description-similarity evaluation and reward computation; and GPT-Image-1.5 is used for defect-guided image refinement. We also used general-purpose LLMs for manuscript writing assistance, including language polishing and wording refinement. All technical claims, experimental results, and final text were reviewed and edited by the authors.

\section{Extended Experimental Results}
\label{app:extended_results}

\subsection{SDG Detector Ablation}
\label{app:sdg_ablation}

Table~\ref{tab:ablation} reports the full ablation results summarized in Section~\ref{sec:main_results}.

\begin{table*}[h]
\centering
\caption{Ablation study on SDG-30K test set. Table header follows Table~\ref{tab:main_art_mis_results}. ``--'' indicates the component is ablated and the corresponding metric is undefined. Ablation groups are separated by dashed lines: training stage, CoT steps, output component, and architecture/augmentation.}
\label{tab:ablation}
\small
\setlength{\tabcolsep}{3pt}
\renewcommand{\arraystretch}{1.15}
\resizebox{\linewidth}{!}{%
\begin{tabular}{l|cccccc|cccccc}
\toprule
& \multicolumn{6}{c|}{\emph{Artifact}} & \multicolumn{6}{c}{\emph{Misalignment}} \\
\cmidrule(lr){2-7}\cmidrule(lr){8-13}
& \multicolumn{2}{c}{Image-Level} & \multicolumn{4}{c|}{Defect-Level} & \multicolumn{2}{c}{Image-Level} & \multicolumn{4}{c}{Defect-Level} \\
\cmidrule(lr){2-3}\cmidrule(lr){4-7}\cmidrule(lr){8-9}\cmidrule(lr){10-13}
& All & $GT{=}0$ & \multicolumn{4}{c|}{$GT{>}0$} & All & $GT{=}0$ & \multicolumn{4}{c}{$GT{>}0$} \\
Variant & DetTypeF1 & ClnAcc & BoxF1@0.1 & BoxF1@0.5 & DescCos@0.1 & ImpAcc@0.1 & DetTypeF1 & ClnAcc & BoxF1@0.1 & BoxF1@0.5 & DescCos@0.1 & ImpAcc@0.1 \\
\midrule
SFT & 0.776 & 0.697 & 0.402 & 0.255 & 0.904 & 0.883 & 0.636 & 0.799 & 0.499 & 0.376 & 0.893 & 0.892 \\
GRPO & 0.772 & 0.560 & 0.404 & 0.263 & 0.904 & 0.887 & 0.675 & 0.732 & 0.511 & 0.387 & 0.888 & 0.893 \\
\cdashline{1-13}
SFT w/o step 1 & 0.719 & 0.793 & 0.357 & 0.218 & 0.894 & 0.880 & 0.597 & 0.798 & 0.436 & 0.329 & 0.875 & 0.892 \\
SFT w/o step 3 & 0.731 & 0.537 & 0.387 & 0.253 & 0.897 & 0.884 & 0.496 & 0.840 & 0.364 & 0.281 & 0.846 & 0.878 \\
GRPO (no think) & 0.776 & 0.640 & 0.417 & 0.288 & 0.899 & 0.881 & 0.640 & 0.677 & 0.487 & 0.352 & 0.879 & 0.883 \\
\cdashline{1-13}
SFT w/o desc & 0.757 & 0.718 & 0.386 & 0.250 & -- & 0.884 & 0.609 & 0.782 & 0.468 & 0.358 & -- & 0.894 \\
SFT w/o importance & 0.761 & 0.745 & 0.402 & 0.254 & 0.898 & -- & 0.594 & 0.793 & 0.448 & 0.357 & 0.870 & -- \\
\cdashline{1-13}
SFT unfreeze ViT & 0.659 & 0.638 & 0.329 & 0.203 & 0.894 & 0.880 & 0.584 & 0.799 & 0.427 & 0.305 & 0.856 & 0.880 \\
SFT w/o jitter & 0.751 & 0.684 & 0.396 & 0.253 & 0.898 & 0.884 & 0.611 & 0.786 & 0.462 & 0.360 & 0.878 & 0.894 \\
\bottomrule
\end{tabular}%
}
\vspace{-14pt}
\end{table*}

\subsection{Extended Comparison with ImageDoctor on SDG-30K}
\label{app:sdg_extended_comparison}

\begin{figure}[H]
\centering
\includegraphics[width=0.92\textwidth]{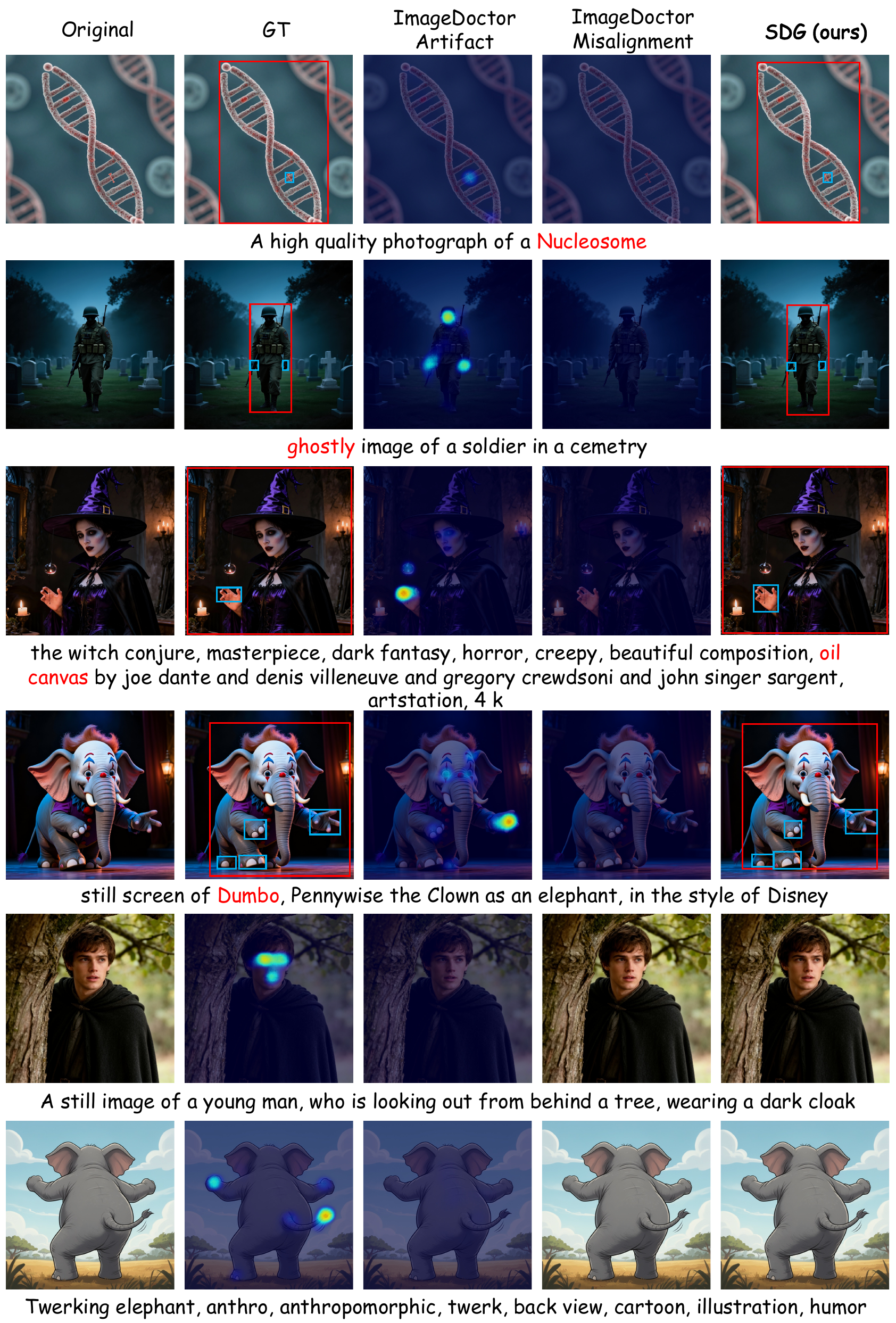}
\caption{Extended qualitative comparison on SDG-30K. Columns: original image, ground-truth SDG annotations, ImageDoctor artifact heatmap, ImageDoctor misalignment heatmap, and SDG (ours) predictions. Red text in the captions highlights the prompt span corresponding to the misalignment.}
\label{fig:sdg_qualitative_extended}
\end{figure}

Figure~\ref{fig:sdg_qualitative_extended} extends the comparison in Figure~\ref{fig:sdg_qualitative} with six additional cases covering both \textit{artifact} and \textit{misalignment} defects, as well as images where the targeted defect type is absent. For ImageDoctor we show the artifact and misalignment heatmap heads separately; for SDG we overlay the predicted bounding boxes with per-instance labels. SDG consistently grounds prompt-conditioned misalignments (e.g., ``Nucleosome'' depicted as a double helix, ``Dumbo'' drawn without clown makeup) that ImageDoctor's misalignment head misses, while avoiding the spurious face/hand activations that ImageDoctor's artifact head produces on clean regions (rows~5--6).

\subsection{Full SDG Output Example}
\label{app:full_sdg_output}

We provide the complete SDG model output for the example shown in Figure~\ref{fig:heatmap_vs_defectset}. The model receives the generated image and the prompt, and produces a structured reasoning trace (\texttt{<think>}) followed by a JSON defect set (\texttt{<answer>}).

\begin{tcolorbox}[breakable, colback=green!3!white, colframe=green!50!black, title=SDG Full Output for Figure~\ref{fig:heatmap_vs_defectset}, left=3pt, right=3pt, top=2pt, bottom=2pt]
\small\texttt{\textbf{Prompt:} Psychedelic smoke, explosion, fire twirling, backlit, twisting, curled, petite American ballerina, wearing ballerina sparkling lace tutu, riding long glowing neon skateboard, star jump upside down, 8K, HD, highly detailed, rendered in octane}

\medskip
\small\texttt{\xmltag{<think>}}

\smallskip
\small\phead{\#\#\# Step 1: Caption Understanding}\\
\small\texttt{- The caption describes a ``petite American ballerina'' in a ``sparkling lace tutu'' riding a ``long glowing neon skateboard''.}\\
\small\texttt{- Key action: ``star jump upside down''.}\\
\small\texttt{- Setting/Style: ``psychedelic smoke'', ``explosion'', ``fire twirling'', ``backlit'', ``twisting'', ``curled'', ``8K'', ``highly detailed'', ``rendered in octane''.}

\smallskip
\small\phead{\#\#\# Step 2: Visual Analysis \& Defect Spotting (Issue Summary)}\\
\small\texttt{- \textbf{Misalignment (Action/Pose):} The most prominent issue is the ballerina's pose. The caption explicitly requests a ``star jump upside down''. However, the generated image shows the ballerina standing upright on the skateboard, balancing on one leg with the other extended. She is neither upside down nor performing a star jump.}\\
\small\texttt{- \textbf{Artifact (Anatomy -- Hands):} Both hands show generation artifacts. The hand raised high in the air (image-right) appears to have fused or indistinct fingers. The hand extended to the side (image-left) also lacks clear finger definition and looks somewhat malformed or ``melted''.}\\
\small\texttt{- \textbf{Artifact (Anatomy -- Face):} The facial features, while detailed, show some subtle distortion typical of AI generation, particularly in the alignment of the eyes and mouth, which appear slightly uncanny or warped.}

\smallskip
\small\phead{\#\#\# Step 3: Localization (Box-by-Box Grounding)}\\
\small\texttt{- \textbf{Box 1 (Artifact):} The defect is located on the raised hand on the image-right side; it is a small localized region where the fingers are visible but appear fused and lack proper anatomical separation.}\\
\small\texttt{- \textbf{Box 2 (Artifact):} The defect is on the extended hand on the image-left side; it is a small localized region at the end of the arm where the fingers are blurry and indistinct, failing to resolve into a clear hand shape.}\\
\small\texttt{- \textbf{Box 3 (Artifact):} The defect is centered on the ballerina's face; it is a part-sized region where the facial geometry (eyes, nose, mouth) appears slightly distorted and lacks the crisp symmetry expected of a high-quality render.}\\
\small\texttt{- \textbf{Box 4 (Misalignment):} The defect encompasses the entire figure of the ballerina; it is a large area covering the subject who is standing upright on the board, contradicting the prompt's instruction for an ``upside down'' and ``star jump'' pose.}

\smallskip
\small\texttt{\xmltag{</think>}}

\medskip
\small\texttt{\xmltag{<answer>}}
\begin{verbatim}
[
  {
    "box_2d": [600, 48, 641, 100],
    "label": "artifact",
    "description": "On the raised image-right hand, the fingers
      appear fused and indistinct, lacking clear anatomical
      separation and joint definition, resulting in a distorted,
      unnatural shape.",
    "importance": 22
  },
  {
    "box_2d": [223, 345, 285, 385],
    "label": "artifact",
    "description": "On the extended image-left hand, the fingers
      are malformed and blurry, showing a lack of distinct digits
      and a melted appearance rather than a properly articulated
      hand.",
    "importance": 22
  },
  {
    "box_2d": [474, 220, 530, 281],
    "label": "artifact",
    "description": "The facial features in the upper center show
      subtle distortion, with the eyes and mouth appearing slightly
      warped or asymmetrical, failing to maintain a consistent
      realistic geometry.",
    "importance": 32
  },
  {
    "box_2d": [223, 48, 657, 840],
    "label": "misalignment",
    "description": "The ballerina is depicted standing upright on
      the skateboard, which directly contradicts the caption's
      explicit requirement for an 'upside down' star jump action.",
    "importance": 95
  }
]
\end{verbatim}
\small\texttt{\xmltag{</answer>}}
\end{tcolorbox}

\subsection{Extended BoxFlow-GRPO Qualitative Comparison}
\label{app:boxflow_extended}

\begin{figure}[H]
\centering
\vspace{-6pt}
\includegraphics[width=0.92\textwidth]{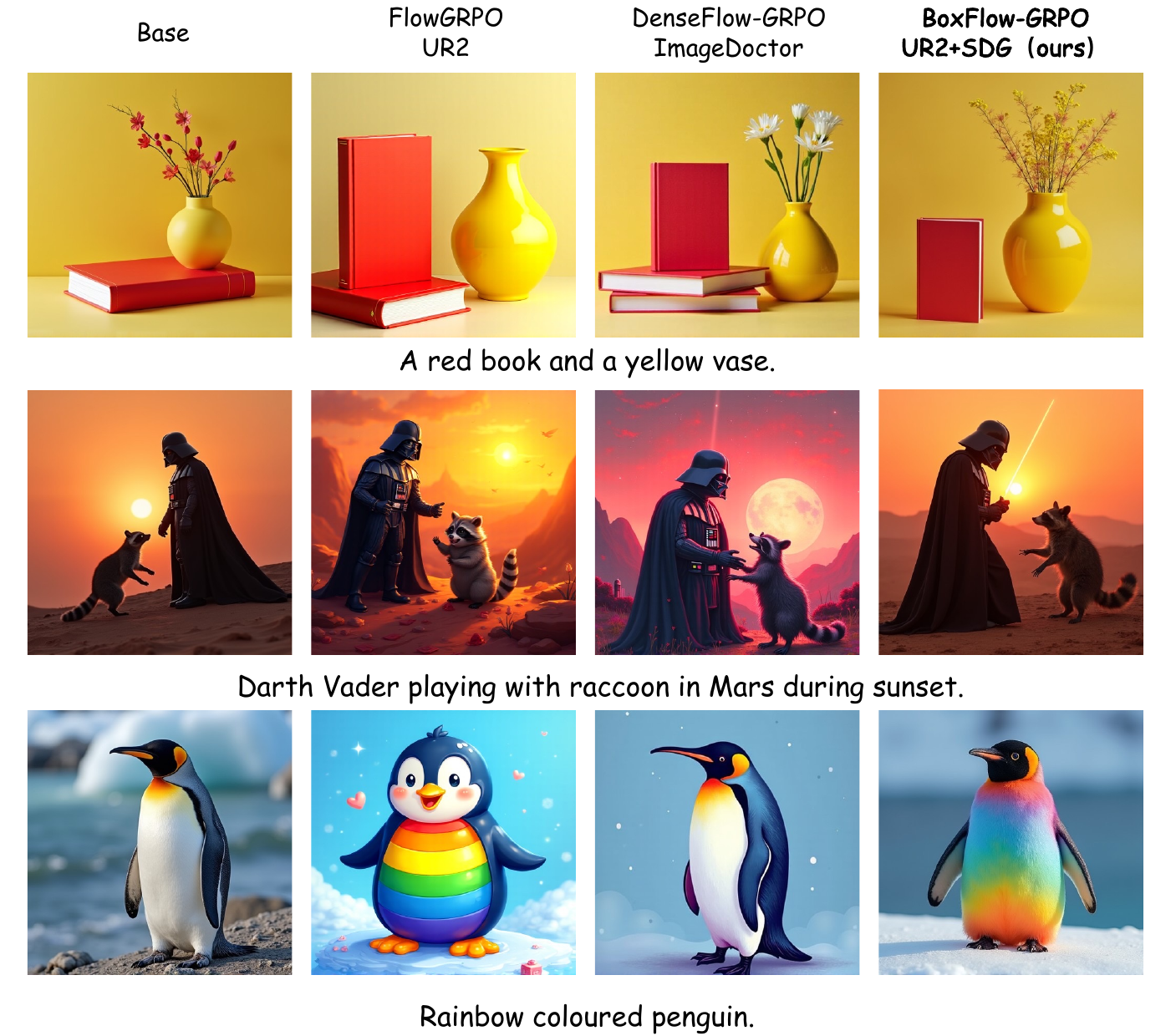}
\vspace{-8pt}
\caption{Extended qualitative comparison of BoxFlow-GRPO on DrawBench prompts. Columns: Base (FLUX.1-dev), FlowGRPO with UR2 reward, DenseFlow-GRPO with ImageDoctor heatmap reward, and BoxFlow-GRPO with UR2+SDG structured reward (ours).}
\label{fig:denseflow_vis_extended}
\vspace{-10pt}
\end{figure}

Figure~\ref{fig:denseflow_vis_extended} extends the qualitative comparison in Figure~\ref{fig:denseflow_vis} with additional DrawBench prompts. Compared to FlowGRPO (UR2) and DenseFlow-GRPO (ImageDoctor), BoxFlow-GRPO (UR2+SDG) better respects fine-grained prompt attributes (e.g., correct color binding in ``a red book and a yellow vase'', coherent subject composition in ``Darth Vader playing with raccoon'') while preserving photographic realism, avoiding the illustration/anime drift that baseline RL variants exhibit.

\subsection{Extended Defect-Guided Refinement Results}
\label{app:reflection_extended}

\begin{figure}[H]
\centering
\vspace{-6pt}
\includegraphics[width=0.92\textwidth]{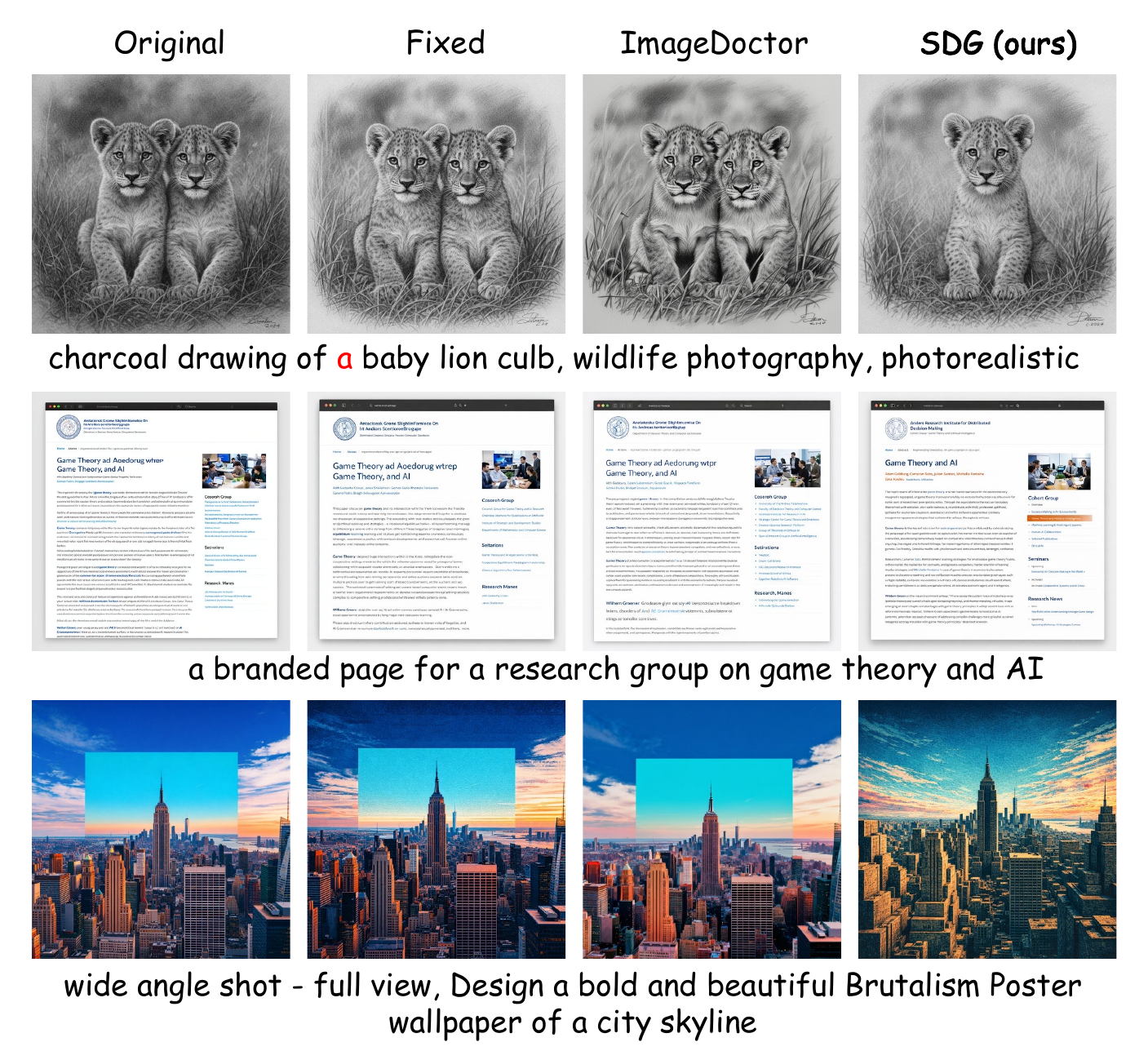}
\vspace{-8pt}
\caption{Extended qualitative comparison of defect-guided image refinement via GPT-Image-1.5. Columns: original image, Fixed caption-only editing, ImageDoctor heatmap/text-feedback editing, and SDG (ours) box-structured-feedback editing.}
\label{fig:reflection_extended}
\vspace{-10pt}
\end{figure}

Figure~\ref{fig:reflection_extended} extends the refinement comparison in Figure~\ref{fig:reflection} with additional GPT-Image-1.5 editing cases. Across these examples, SDG feedback provides localized, instance-level guidance that helps the editor make targeted corrections, such as removing an extra lion cub when the prompt asks for a single cub, identifying an incorrect webpage title and replacing it with the correct title, and eliminating a large poster-like obstruction from a city-skyline image.

\subsection{Validation of Importance Scores Against Human Judgments}
\label{app:importance_human_validation}

Importance scores provide supervision for the SDG detector and contribute to both the GRPO reward and the importance-weighted spatial penalties in BoxFlow-GRPO. To assess whether Gemini-derived importance scores reflect human judgments of perceptual severity, we randomly sampled 500 distinct defect boxes and assigned 250 non-overlapping boxes to each of two human annotators. Each annotator reassessed severity using the same five-tier rubric without seeing the Gemini scores. Thus, each box received one human rating, which we compared with the corresponding Gemini assessment.

\begin{table}[h]
\centering
\caption{Agreement between human severity judgments and Gemini-derived importance scores on 500 distinct defect boxes. Two annotators each assessed 250 non-overlapping boxes while blinded to Gemini scores.}
\label{tab:importance_human_validation}
\begin{tabular}{lc}
\toprule
Human calibration metric & Result \\
\midrule
Exact-tier agreement & 64.8\% \\
Agreement within one adjacent tier & 97.2\% \\
Spearman rank correlation & 0.786 \\
\bottomrule
\end{tabular}
\end{table}

Table~\ref{tab:importance_human_validation} summarizes the results. Exact-tier agreement requires the same severity tier, while adjacent-tier agreement allows a difference of at most one tier. Spearman correlation of 0.786 indicates strong agreement in severity ordering; 97.2\% of assessments differ by at most one tier. These results support importance as a proxy for perceptual severity on the sampled defects. The disjoint assignments measure human--Gemini agreement, not inter-annotator reliability.

\subsection{Area-Normalized Penalty Coefficients and Sensitivity Analysis}
\label{app:coefficient_sensitivity}

\paragraph{Area-normalization rationale.}
The ratio between $c_{\mathrm{art}}$ and $c_{\mathrm{mis}}$ is motivated by the different spatial extents of the two defect types. An average misalignment box covers approximately $9.90\times$ the area of an artifact box. Since BoxFlow-GRPO applies penalties at covered latent locations, equal coefficients would assign a larger total spatial penalty to a larger box at comparable importance. Writing $\bar{A}_{\mathrm{art}}$ and $\bar{A}_{\mathrm{mis}}$ for the average box areas, approximate area normalization gives
\begin{equation}
\frac{c_{\mathrm{mis}}}{c_{\mathrm{art}}}
\approx \frac{\bar{A}_{\mathrm{art}}}{\bar{A}_{\mathrm{mis}}}
\approx \frac{1}{9.90}.
\end{equation}
We therefore use $c_{\mathrm{art}}{=}0.5$ and $c_{\mathrm{mis}}{=}0.05$. Their area-scaled contributions have ratio $(0.05\times9.90)/0.5=0.99$, approximately balancing spatial penalty mass. This is an area-based normalization rationale rather than an exact equality of realized penalties, which also depend on importance weights and box overlap.

\paragraph{Sensitivity analysis.}
We fixed $c_{\mathrm{art}}{=}0.5$ and varied $c_{\mathrm{mis}}\in\{0.25,0.05,0.025\}$. All three variants were trained for 500 steps and evaluated on DrawBench, using the same evaluation set as Table~\ref{tab:ur2_box_results}. Table~\ref{tab:coefficient_sensitivity} reports five evaluation metrics and their mean relative change from Base (Avg.), following Table~\ref{tab:ur2_box_results}.

\begin{table}[ht]
\centering
\caption{Sensitivity to the artifact/misalignment penalty coefficients. All trained variants use 500 steps and are evaluated on DrawBench. Avg. is the mean relative improvement over Base across the five metrics. The default setting is bolded.}
\label{tab:coefficient_sensitivity}
\small
\setlength{\tabcolsep}{5pt}
\begin{tabular}{lrrrrrr}
\toprule
$c_{\mathrm{art}} / c_{\mathrm{mis}}$ & PickScore & CLIPScore & HPSv3 & DeQA & $P(\mathrm{real})$ & Avg. \\
\midrule
Base & 22.84 & 0.912 & 11.75 & 0.876 & 0.211 & --- \\
$0.5 / 0.25$ & 22.83 & 0.910 & 11.87 & 0.874 & 0.217 & $+0.7\%$ \\
$\mathbf{0.5 / 0.05}$ & \textbf{22.91} & \textbf{0.915} & \textbf{12.14} & \textbf{0.877} & \textbf{0.228} & $\mathbf{+2.4\%}$ \\
$0.5 / 0.025$ & 22.86 & 0.910 & 11.92 & 0.861 & 0.226 & $+1.3\%$ \\
\bottomrule
\end{tabular}
\end{table}

All three settings improve the average relative to Base. The area-normalized setting $(0.5,0.05)$ achieves the largest average improvement ($+2.4\%$) and is the only tested setting that improves all five metrics relative to Base. Increasing $c_{\mathrm{mis}}$ to $0.25$ slightly reduces PickScore, CLIPScore, and DeQA relative to Base; decreasing it to $0.025$ reduces CLIPScore and DeQA. These results support $(0.5,0.05)$ as the most balanced operating point among the tested settings.

\subsection{Closing the Diagnosis-to-Refinement Loop with Paired Re-diagnosis}
\label{app:refinement_rediagnosis}

\paragraph{Evaluation protocol.}
To test whether refinement reduces defects as assessed by the detector itself, we applied the same frozen SDG detector to paired images before and after editing. We compared SDG-guided refinement, ImageDoctor-guided refinement, and direct caption-only editing (the \textbf{Fixed} baseline in Section~\ref{sec:reflection_results}). We report the percentage reductions in detected defect boxes and in their total importance relative to the corresponding pre-edit images. Total importance sums the importance scores of detected defects, capturing their combined predicted severity rather than only their count.

\begin{table}[ht]
\centering
\caption{Paired re-diagnosis before and after refinement with the same frozen SDG detector. Larger reductions indicate fewer detected defects and lower total predicted importance after editing.}
\label{tab:refinement_rediagnosis}
\begin{tabular}{lcc}
\toprule
Editing method & Box reduction & Importance reduction \\
\midrule
\textbf{SDG (ours)} & \textbf{37.55\%} & \textbf{42.37\%} \\
ImageDoctor & 25.81\% & 30.36\% \\
Fixed (direct edit) & 30.71\% & 34.17\% \\
\bottomrule
\end{tabular}
\end{table}

\paragraph{Results and interpretation.}
All three editing methods reduce both detected defect count and total importance (Table~\ref{tab:refinement_rediagnosis}). SDG-guided refinement achieves the largest reduction in both measures: 37.55\% fewer boxes and 42.37\% lower total importance. Thus, its improvements extend beyond reducing the number of detected regions to reducing their combined predicted severity. With the detector held fixed, these changes support sensitivity to the corrected image content, although aggregate reductions alone do not rule out location-specific biases such as repeated activation on hands or faces.

This paired re-diagnosis complements the human preference evaluation in Table~\ref{tab:gsb_all_samples}. Together, the human preference results and reductions in re-diagnosed defects provide evidence that structured diagnosis translates into effective image refinement, closing the diagnosis-to-refinement loop at the detector level.

\subsection{Generalization to the Qwen-Image Base Model}
\label{app:qwen_image_results}

To evaluate the method on another base model, we trained BoxFlow-GRPO, Flow-GRPO with UR2, and DenseFlow-GRPO with ImageDoctor on Qwen-Image for 1,200 steps each. We use the same evaluation dataset as the FLUX.1-dev experiments reported in Table~\ref{tab:ur2_box_results}. Table~\ref{tab:qwen_image_results} reports the five metrics used in Table~\ref{tab:ur2_box_results}, with relative changes computed against the Qwen-Image Base model. Avg is the arithmetic mean of the five per-metric relative changes, rather than an average of their raw scores.

\begin{table}[ht]
\centering
\caption{Downstream performance on Qwen-Image after 1,200 training steps for each RL variant. Parentheses show relative changes from Qwen-Image Base (\textcolor{green!60!black}{green}: improvement; \textcolor{red!80!black}{red}: regression); Avg averages these changes across the five metrics. Best values are bold and second-best values are underlined, including Base. BoxFlow-GRPO achieves the best Avg.}
\label{tab:qwen_image_results}
\footnotesize
\setlength{\tabcolsep}{3pt}
\renewcommand{\arraystretch}{1.2}
\resizebox{\linewidth}{!}{%
\begin{tabular}{ll|cccccc}
\toprule
Method & Reward & PickScore & CLIPScore & HPSv3 & DeQA & $P(\mathrm{real})$ & Avg \\
\midrule
Base & -- & 23.03 & \textbf{0.969} & 10.90 & 0.843 & \underline{0.230} & -- \\
\midrule
Flow-GRPO & UR2 & \textbf{23.62}\,\textcolor{green!60!black}{\tiny(+2.6\%)} & 0.923\,\textcolor{red!80!black}{\tiny(-4.7\%)} & \textbf{12.65}\,\textcolor{green!60!black}{\tiny(+16.1\%)} & \textbf{0.868}\,\textcolor{green!60!black}{\tiny(+3.0\%)} & 0.177\,\textcolor{red!80!black}{\tiny(-23.0\%)} & \underline{\textcolor{red!80!black}{-1.2\%}} \\
\midrule
DenseFlow-GRPO & ImageDoctor & 23.14\,\textcolor{green!60!black}{\tiny(+0.5\%)} & 0.941\,\textcolor{red!80!black}{\tiny(-2.9\%)} & \underline{11.07}\,\textcolor{green!60!black}{\tiny(+1.6\%)} & 0.764\,\textcolor{red!80!black}{\tiny(-9.4\%)} & 0.195\,\textcolor{red!80!black}{\tiny(-15.2\%)} & \textcolor{red!80!black}{-5.1\%} \\
\midrule
\textbf{BoxFlow-GRPO} (ours) & UR2+SDG & \underline{23.16}\,\textcolor{green!60!black}{\tiny(+0.6\%)} & \underline{0.965}\,\textcolor{red!80!black}{\tiny(-0.4\%)} & 11.05\,\textcolor{green!60!black}{\tiny(+1.4\%)} & \underline{0.852}\,\textcolor{green!60!black}{\tiny(+1.1\%)} & \textbf{0.237}\,\textcolor{green!60!black}{\tiny(+3.0\%)} & \textbf{\textcolor{green!60!black}{+1.1\%}} \\
\bottomrule
\end{tabular}%
}
\end{table}

BoxFlow-GRPO again achieves the best average relative improvement ($+1.1\%$), compared with $-1.2\%$ for Flow-GRPO (UR2) and $-5.1\%$ for DenseFlow-GRPO (ImageDoctor). It improves PickScore, HPSv3, DeQA, and $P(\mathrm{real})$, with a small CLIPScore decrease ($-0.4\%$). Flow-GRPO (UR2) obtains the highest PickScore, HPSv3, and DeQA, but its $P(\mathrm{real})$ drops from 0.230 to 0.177 ($-23.0\%$). BoxFlow-GRPO instead increases $P(\mathrm{real})$ to 0.237 ($+3.0\%$), yielding a more balanced outcome across the five metrics. These results extend the favorable average performance observed on FLUX.1-dev to Qwen-Image.

\section{Limitations and Broader Impact}
\label{app:limitations_broader_impact}

\subsection{Limitations}

SDG currently focuses on two defect types, \textit{artifact} and \textit{misalignment}. Other quality dimensions, such as aesthetics, style, composition, safety, and cultural appropriateness, fall outside the current label space and would require additional annotation guidelines and evaluation metrics. The dataset is constructed from four contemporary T2I generators and Pick-a-Pic prompts; although this provides broad coverage, performance may change for other generators, domains, resolutions, or prompt distributions.

Our importance scores are distilled from Gemini 3 Pro under a fixed rubric. Human calibration on 500 defect boxes supports their agreement with human severity judgments (Appendix~\ref{app:importance_human_validation}), but these scores may still differ from human preferences and inherit biases from the teacher model. The SDG detector can also miss subtle defects, hallucinate defects in clean locations, or produce boxes that are too coarse for very small or highly diffuse failures. Finally, BoxFlow-GRPO assumes that box-derived penalties can be meaningfully projected onto latent spatial locations; this approximation may be less reliable when the latent grid does not align cleanly with visible defect locations.

\subsection{Broader Impact}

Structured defect feedback can make T2I systems more interpretable by exposing localized failure modes, supporting dataset auditing, and enabling targeted image refinement. These properties can help users diagnose generation errors rather than relying only on scalar preference scores.

The same capabilities also carry risks. Better diagnosis and refinement may improve the realism of synthetic images, which could be misused for deceptive or harmful content. Our public release includes code, model weights, and SDG-30K. The released assets should preserve annotation provenance, document intended use, and include safeguards for controlled release where appropriate. Because SDG can make localized judgments about generated people or scenes, downstream deployments should also consider fairness and bias in the underlying prompts, generators, annotators, and teacher models.

\end{document}